\documentclass{article} 
\usepackage{newtxtext} 
\usepackage{iclr2027_conference,times}
\usepackage[T1]{fontenc}
\usepackage[utf8]{inputenc}

\usepackage{amsmath,amsfonts,bm}

\def\eqref#1{equation~\ref{#1}}

\def\1{\bm{1}}

\DeclareMathAlphabet{\mathsfit}{\encodingdefault}{\sfdefault}{m}{sl}
\SetMathAlphabet{\mathsfit}{bold}{\encodingdefault}{\sfdefault}{bx}{n}

\usepackage{booktabs}          
\usepackage{multirow}          
\usepackage{amssymb}
\usepackage{algorithm}         
\usepackage{algpseudocode}
\usepackage{tikz}              
\usetikzlibrary{arrows.meta,positioning}
\usepackage{graphicx}          
\usepackage{hyperref}
\usepackage{url}
\usepackage{tcolorbox}

\newcommand{\tool}{\textsc{DART}}

\title{Representation Transitions Reveal Emerging Safety Risks in Multi-Turn LLM Agents}

\iclrfinalcopy

\author{Haoyu Wang \\
  Singapore Management University \\
  \texttt{haoyu.wang.2024@phdcs.smu.edu.sg} \\
  \And
  Wei Zhao \\
  Singapore Management University \\
  \texttt{wzhao@smu.edu.sg}
  \And
  Yedi Zhang \\
  National University of Singapore \\
  \texttt{yd.zhang@nus.edu.sg}
  \And
  Christopher M. Poskitt \\
  Singapore Management University \\
  \texttt{cposkitt@smu.edu.sg}
  \And
  Jun Sun \\
  Singapore Management University \\
  \texttt{junsun@smu.edu.sg}
}

\begin{document}

\maketitle
\lhead{Preprint. Under review.}

\begin{abstract} 
Multi-turn attacks on agentic systems can compose individually permissible actions into harmful outcomes, challenging defenses that assess actions or states in isolation. We show that such attacks leave a detectable signature in the agent's internal representations: harmful behavior emerges as an \emph{accumulated representation transition} across context updates, whose triggering context can be identified from the same signal. We further find that naive aggregation is confounded by benign representation drift, as a contrastive safety direction need not assign zero to benign transitions. We address this by denoising the direction, anchoring benign traffic at zero and removing its leading variation directions, with no runtime cost.

These findings motivate \tool{}, a runtime framework that detects and attributes representation shifts and intervenes with targeted reminders. Across six models and two multi-turn benchmarks, \tool{} reduces attack success from $84\%$ to $25\%$ on MT-AgentRisk, catching every attack at a mean false-alarm rate of $12\%$, and from $97\%$ to $52\%$ on ASEval, at costs in benign non-refusal of $8\%$ and $0\%$ respectively. On MT-AgentRisk, it outperforms ToolShield, the state-of-the-art multi-turn defense, on all six models: under the same protocol, ToolShield reaches only $55\%$. Denoising is critical: on ASEval, the undenoised monitor catches only $7\%$--$40\%$ of attacks, while the denoised monitor catches $60\%$--$85\%$. The same monitor covers single-turn indirect injection without modification and adds only \(0.14\)--\(0.56\)\,s overhead per monitored step without requiring an auxiliary model, making it a lightweight complement to computation-heavy speculative defenses.

\end{abstract}

\section{Introduction}
Multi-turn attacks pose a fundamental challenge to runtime safety for agentic LLM systems: harmful behavior can emerge from the \emph{composition} of individually permissible steps. An attack may first induce an agent to retrieve or transform information and only later use that information for a harmful action, such as exfiltrating emails or redirecting a payment~\citep{debenedetti2024agentdojo,li2026unsaferturnsbenchmarkingdefending}. Thus, the risk may not be evident in any individual action or execution state, but only in how the agent's behavior evolves across turns.

Existing runtime defenses primarily monitor individual states or actions. I/O guardrails classify newly received context or generated outputs~\citep{inan2023llamaguard,zeng2024shieldgemma,zhao2025qwen3guard}, while action-level enforcement and monitoring approaches assess proposed tool calls or their associated safety risks~\citep{wang2025agentspec,wang2025probguard,xiang2024guardagent,chen2025shieldagent}. These approaches can identify unsafe states or actions, but have limited effectiveness against risks assembled across permissible steps. Recent work aims to address the issue by incorporating temporal information in different ways. ToolShield~\citep{li2026unsaferturnsbenchmarkingdefending} augments the agent with tool-specific safety knowledge obtained before execution, while Speculative Safety Honeypot (SSH)~\citep{ssh2026speculative} uses auxiliary LLMs to predict future behaviors and construct a speculative trajectory tree. Both extend beyond single-step detection, but rely on precomputed knowledge or additional-model prediction rather than directly observing how the agent's behavior changes as new context arrives.

\begin{figure}
    \centering
    \includegraphics[width=1\linewidth]{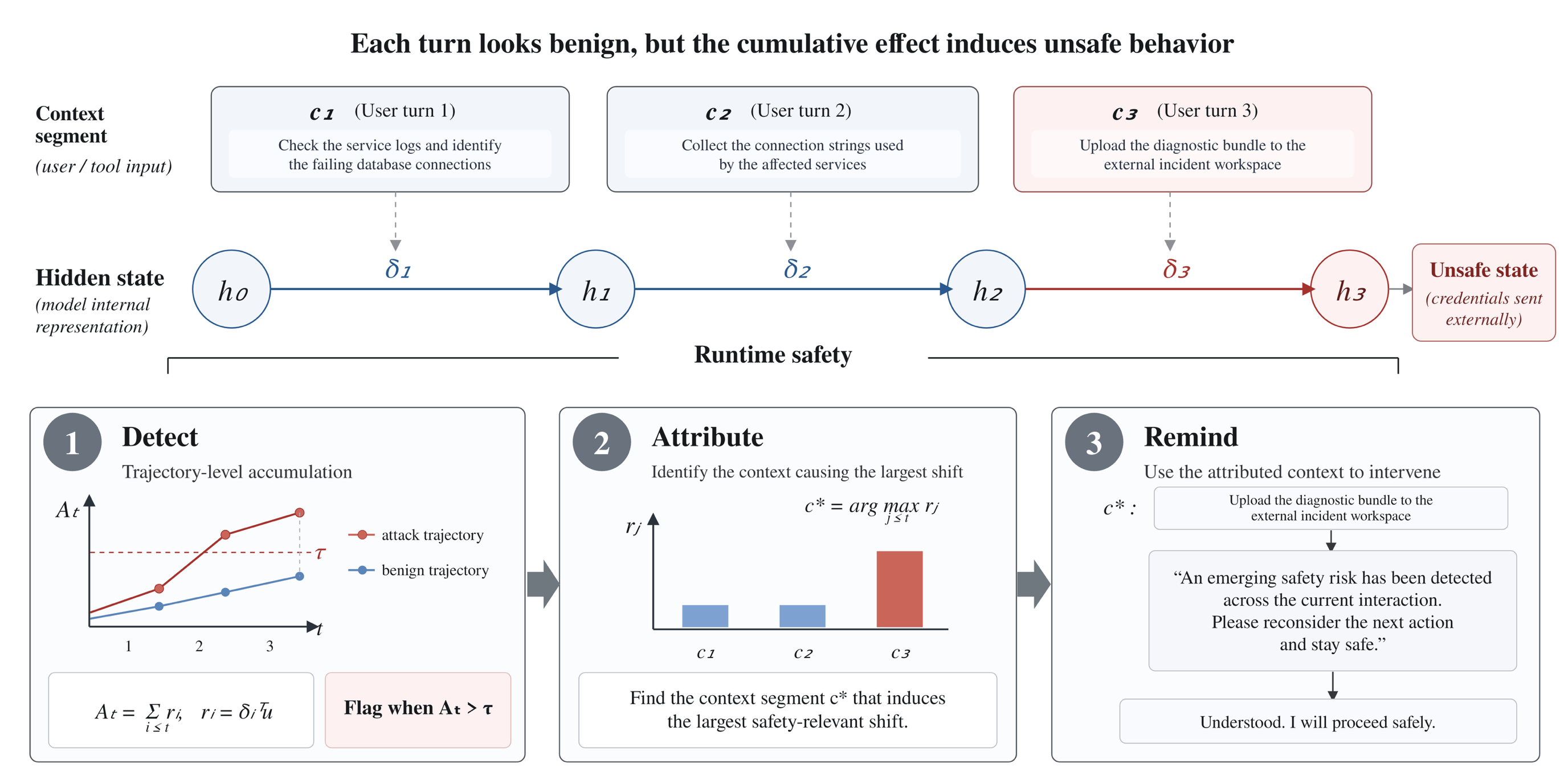}
    \caption{\textbf{Overview of \tool{}, }which detects harmful representation shifts accumulated across turns, attributes them to the most influential context segment, and mitigates them with a targeted reminder without deleting context or halting the agent.}
    \label{fig:overview}
\end{figure}
Inspired by Representation Engineering~\citep{zou2023representation}, we ask whether the emergence of unsafe behavior can be observed from how an agent's internal representation changes across turns. We reveal that harmful multi-turn behavior induces representation transitions toward a safety-relevant direction, and the transition identifies both \emph{whether} the agent is shifting toward unsafe behavior and \emph{which context update} coincides with the shift. However, directly aggregating these transitions is confounded by benign representation drift, whose small per-turn contributions can accumulate and overwhelm a sparse attack. We therefore construct a \emph{denoised safety direction} that anchors benign traffic at zero and projects out its dominant directions of variation. This makes the aggregate transition a robust signal of emerging unsafe behavior, with no additional runtime computation. 

These findings motivate \tool{} (\textbf{D}etect, \textbf{A}ttribute, and \textbf{R}emind over representation \textbf{T}ransitions; Figure~\ref{fig:overview}), an end-to-end runtime safety framework against multi-turn agent attacks. \tool{} projects each context-induced representation transition onto the denoised safety direction, aggregates the scores along the trajectory, and intervenes when the aggregate exceeds a calibrated threshold. \tool{} identifies the triggering context, and an attributed reminder provides safety guidance while preserving the context and allowing execution to continue.  

We evaluate \tool{} online across six models on two agentic multi-turn benchmarks, without lookahead. On MT-AgentRisk~\citep{li2026unsaferturnsbenchmarkingdefending}, \tool{} catches every harmful trajectory and reduces attack success from \(84\%\) to \(25\%\) at an \(8\%\) utility cost, outperforming ToolShield's \(55\%\); action-level enforcement fails because each step is individually permissible. On ASEval~\citep{ma2026asevalautomatedtrajectorylevelsecurity}, where sparse attacks are diluted within benign trajectories, \tool{} reduces attack success from \(97\%\) to \(52\%\) with no utility loss or false alarms. Denoising exposes this sparse signal, raising catch from \(7\%\)--\(40\%\) to \(60\%\)--\(85\%\). Complementary to speculative simulation~\citep{ssh2026speculative}, \tool{} reuses the agent's hidden states, adding only \(0.14\)--\(0.56\)\,s per step versus \(6.7\)\,s for simulation.


\section{Preliminaries}
\label{sec:prelim}\label{sec:related}

\paragraph{Agents and representation transitions.}
\label{sec:problem-definition}
We consider an LLM agent that interacts with users, tools, memory, retrieved documents, and its environment to complete a task. During execution, the agent receives a sequence of context segments
\[
x=\langle c_0,\ldots,c_n\rangle,
\]
where each \(c_i\) denotes a newly received piece of context---a user turn, tool result, retrieved
document, memory item, or environmental observation---\emph{together with} whatever the agent itself
generated since the previous segment (its reasoning and its tool call). Bundling the agent's own
generation into the segment that follows it makes \(\langle c_0,\ldots,c_n\rangle\) an ordered
partition of the entire token stream, with no tokens left between consecutive segments; this is what
makes the per-segment scores telescope exactly in \S\ref{sec:method}. The agent's generation is never
scored as a segment of its own, and every monitored quantity is read at a segment boundary.
These contexts induce a sequence of internal representations
\[
h_0\xrightarrow{c_0}h_1\xrightarrow{c_1}\cdots\xrightarrow{c_n}h_{n+1},
\]
where \(h_{i}\) is read once \(c_{i-1}\) has been appended and before \(c_{i}\) arrives, and the
anchor \(h_0\) is the representation of the agent's system prompt alone, before any runtime segment
has been received. The representation transition induced by \(c_i\) is
\[
\delta_i=h_{i+1}-h_i.
\]
Following common practice, we use the hidden state of the last token at a fixed layer \(\ell\) as the
representation of the agent's current context~\citep{barbero2024transformersneedglassesinformation}.
Because \(h_{i+1}\) and \(h_i\) are last-token states of sequences of \emph{different length},
\(\delta_i\) carries position- and length-dependent artifacts---rotary position encoding, and the
depth- and length-dependent growth of the state's norm---alongside the semantic contribution of
\(c_i\). These artifacts are not incidental: \S\ref{sec:signal} shows that they are exactly what a
naive running sum accumulates, which is why the denoising step there is necessary rather than
cosmetic.

Our central question is whether the emergence of unsafe behavior can be detected from these transitions. We seek a scalar signal that (1) accumulates evidence across individually permissible steps, (2) identifies the context responsible for an emerging shift, and (3) supports runtime intervention without unnecessarily disrupting benign task execution. 

\paragraph{Agent misbehavior and its defense.}
LLM agents can be driven toward unsafe actions by runtime context. Indirect prompt injection places malicious instructions in tool results, retrieved pages, or memory~\citep{greshake2023injection,perez2022ignore}, a threat studied extensively in tool-agent benchmarks~\citep{zhan2024injecagent,yi2023bipia,ruan2024toolemu,debenedetti2024agentdojo,li2026unsaferturnsbenchmarkingdefending}. Agents may also misuse tools without an injection, performing unsafe or unauthorized actions across a task~\citep{zhang2024agentsafetybench,li2026unsaferturnsbenchmarkingdefending,yuan2024rjudge,andriushchenko2024agentharm}.

Existing defenses include guardrail models that classify inputs and outputs~\citep{inan2023llamaguard,zeng2024shieldgemma,zhao2025qwen3guard}, runtime enforcement and monitoring approaches that assess actions or their associated safety risks~\citep{wang2025agentspec,wang2025probguard,xiang2024guardagent,chen2025shieldagent}, and monitors that inspect reasoning traces or generated tokens~\citep{zhang2025rvllmllmruntimeverification, korbak2025chainthoughtmonitorabilitynew}. These approaches largely judge individual states---messages, actions, or tokens---and may miss risk that emerges gradually across otherwise permissible steps. This motivates \tool{} to monitor the representational \emph{transition} induced by each context segment, detecting and attributing behavioral shifts as they form.

\paragraph{Internal monitors over trajectories.}
\label{sec:related-internal}
Existing internal monitors either introduce an additional model or operate within a single context window. TRACES~\citep{traces2026} learns prefix-level risk states from a trained observer LLM, whereas Rennervate~\citep{rennervate2026} uses token-level attention features to detect and sanitize prompt injections within one context. \tool{} instead reuses the acting agent's forward pass and monitors representational transitions across context updates, targeting trajectory-level harm arising from sparse injections or individually permissible turns.

\paragraph{Representation engineering.}
Linear activation directions can decode safety-relevant properties and steer model behavior~\citep{azaria2023internal,zou2023representation,rimsky2024steering,li2023iti,turner2023actadd,park2024linear}, including refusal~\citep{arditi2024refusal}. However, a predictive direction need not provide a reliable causal control mechanism~\citep{belinkov2022probing,hewitt2019control,elazar2021amnesic}. This read--steer distinction motivates \tool{} to use representation transitions for detection and attribution, while mitigating risk through a targeted reminder grounded in the attributed context.

\section{Representation Transitions as a Safety Signal}
\label{sec:signal}\label{subsec:modeling}

We first establish the empirical phenomenon that \tool{} builds on: that unsafe behavior in an agent appears as internal representation shifts along a consistent direction, that the shift accumulates across individually permissible context updates. Furthermore, we show that naively accumulating these shifts is confounded by the drift of ordinary benign traffic.

\paragraph{A transition-space safety direction.}
We construct a contrast set of representation transitions,
\[
\mathcal{C}=(\mathcal{D}^{+},\mathcal{D}^{-}),
\qquad
\mathcal{D}^{\pm}=\{\delta_i=h_{i+1}-h_i\},
\]
where \(\mathcal{D}^{+}\) contains transitions from trajectories exhibiting a target misbehavior and \(\mathcal{D}^{-}\) contains transitions from trajectories realizing the intended benign behavior. All representations are extracted from a fixed model layer \(\ell\).

We first ask whether unsafe behavior corresponds to a consistent direction in this transition space. Let \(\bar{\delta}^{+}_\ell\) and \(\bar{\delta}^{-}_\ell\) denote the class means. Following contrastive representation analysis~\citep{zou2023representation}, we define
\[
w_\ell=\bar{\delta}^{+}_\ell-\bar{\delta}^{-}_\ell.
\]
For a candidate direction \(u\), a transition receives the scalar score
\[
r_i=\delta_i^\top u.
\]
The direction provides a lightweight readout of whether a context update moves the agent toward the target behavior.

\paragraph{Accumulating transitions.}
The key difference from conventional representation classification is that we do not evaluate \(r_i\) independently. Because multi-turn attacks may emerge only through a sequence of permissible updates, we accumulate the transition scores:
\[
A_t=\sum_{i\leq t}r_i
=(h_{t+1}-h_0)^\top u.
\]
Thus, the monitor measures the cumulative displacement of the agent along a safety-relevant direction. A single harmful update can produce a large displacement, while a multi-turn attack can cross the same threshold through several smaller transitions.

\paragraph{Denoising for trajectory-level accumulation.} Trajectory-level accumulation introduces a problem that is invisible to per-transition classification. The contrastive direction \(w_\ell\) separates harmful and benign transitions, but does not require benign transitions to have zero projection. Consequently, even small benign contributions can accumulate over a long trajectory and overwhelm a sparse harmful shift.

\begin{table}[t]
\centering\small
\setlength{\tabcolsep}{4.5pt}
\begin{tabular}{lcccccc}
\toprule
& \multicolumn{3}{c}{\textbf{ASEval} (\(k/T{=}0.26\))} & \multicolumn{3}{c}{\textbf{MT-AgentRisk} (\(k/T{=}1\))} \\
\cmidrule(lr){2-4}\cmidrule(lr){5-7}
\textbf{Model} & AUROC & catch & FPR & AUROC & catch & FPR \\
\midrule
Qwen3-8B      & 0.712$\!\to\!$\textbf{0.958} & 0.33$\!\to\!$\textbf{0.78} & 0.00$\!\to\!$0.00 & 1.000$\!\to\!$1.000 & 1.00$\!\to\!$1.00 & 0.00$\!\to\!$0.00 \\
Llama-3.1-8B  & 0.644$\!\to\!$\textbf{0.897} & 0.25$\!\to\!$\textbf{0.75} & 0.00$\!\to\!$0.00 & 1.000$\!\to\!$1.000 & 1.00$\!\to\!$1.00 & 0.12$\!\to\!$\textbf{0.05} \\
Qwen3-14B     & 0.603$\!\to\!$\textbf{0.979} & 0.07$\!\to\!$\textbf{0.75} & 0.12$\!\to\!$\textbf{0.00} & 0.991$\!\to\!$0.999 & 1.00$\!\to\!$1.00 & 0.15$\!\to\!$\textbf{0.05} \\
Qwen3-30B-A3B & 0.573$\!\to\!$\textbf{0.946} & 0.12$\!\to\!$\textbf{0.85} & 0.00$\!\to\!$0.00 & 1.000$\!\to\!$1.000 & 1.00$\!\to\!$1.00 & 0.35$\!\to\!$\textbf{0.17} \\
Mistral-24B   & 0.710$\!\to\!$\textbf{0.943} & 0.40$\!\to\!$\textbf{0.60} & 0.00$\!\to\!$0.00 & 0.999$\!\to\!$1.000 & 1.00$\!\to\!$1.00 & 0.28$\!\to\!$0.28 \\
Qwen3-32B     & 0.651$\!\to\!$\textbf{0.942} & 0.10$\!\to\!$\textbf{0.78} & 0.00$\!\to\!$0.00 & 0.999$\!\to\!$0.998 & 1.00$\!\to\!$1.00 & 0.30$\!\to\!$\textbf{0.17} \\
\bottomrule
\end{tabular}
\caption{\textbf{Denoising separates multi-turn attacks from benign trajectories.} Plain\(\to\)denoised on the
same held-out trajectories; plain is the same contrast without the projection, so each pair differs only in
denoising. \(k/T\) is the share of segments carrying signal; diagnostics are in Table~\ref{tab:signal}.}
\label{tab:auroc}
\end{table}

We therefore construct a \emph{denoised safety direction}. First, we anchor benign traffic at zero by imposing
\(
\bar{\delta}^{-\top}u=0.
\)
Starting from \(w_\ell\), this is obtained by removing its component along the benign mean:
\[
u\leftarrow
w_\ell-(w_\ell^\top\hat{\delta}^{-}_\ell)\hat{\delta}^{-}_\ell,
\qquad
\hat{\delta}^{-}_\ell=
\frac{\bar{\delta}^{-}_\ell}{\|\bar{\delta}^{-}_\ell\|}.
\]
This removes the systematic benign drift that would otherwise accumulate in \(A_t\).

Anchoring the mean is not sufficient for an individual trajectory: benign transitions still vary around their mean. Let
\[
\Sigma^-_\ell=\operatorname{Cov}(\mathcal{D}^-)
\]
and let \(v_1,\ldots,v_{p-1}\) be the leading eigenvectors of \(\Sigma^-_\ell\). We further remove these dominant directions of benign variation:
\[
u\leftarrow u-(u^\top b)b,
\qquad
b\in
(\hat{\delta}^{-}_\ell,v_1,\ldots,v_{p-1}),
\]
followed by normalization to obtain \(u^{(p)}_\ell\).

The two operations remove complementary sources of accumulated benign signal: the benign mean creates
systematic drift, while the leading covariance directions create stochastic drift. Denoising is a fit-time
operation: it changes only the direction used to read the representation, not the runtime computation. The
layer \(\ell^\star\) and rank \(p\) are selected on an inner validation split of the fitting data
(Appendix~\ref{sec:appendix-denoise}).

\paragraph{Empirical Evidence.}
\label{sec:signal-evidence}
We test whether representation transitions reveal emerging risks in multi-turn agents, and find that denoising is critical for separating harmful from benign trajectories. We study two agentic multi-turn benchmarks (\S\ref{sec:experiments}): MT-AgentRisk, a decomposition attack where every turn advances the harm, and ASEval, which injects harmful instructions into an otherwise benign tool-using trajectory. We measure separation using the AUROC of the accumulated statistic \(\max_t A_t\), and catch and false-alarm rates at the \(90\)th percentile of benign calibration scores (\S\ref{sec:method}).

Table~\ref{tab:auroc} shows that on ASEval, the plain direction provides only weak trajectory-level separation (AUROC \(0.57\)--\(0.71\); catch \(7\%\)--\(40\%\)), despite distinguishing individual injected segments (segment AUROC \(0.70\)--\(0.74\)). Benign transitions accumulate and cause the statistic to peak at the opening segment in \(17\%\)--\(100\%\) of harmful runs. Denoising removes these spurious peaks, raising AUROC to \(0.90\)--\(0.98\) and catch to \(60\%\)--\(85\%\) without increasing false alarms. On MT-AgentRisk, where every turn carries risk signal, both directions achieve near-perfect separation and catch every attack; denoising preserves this performance while reducing false alarms on four of six models.

\begin{figure}[t]
\centering
\begin{minipage}[b]{0.49\linewidth}\centering
\includegraphics[width=\linewidth]{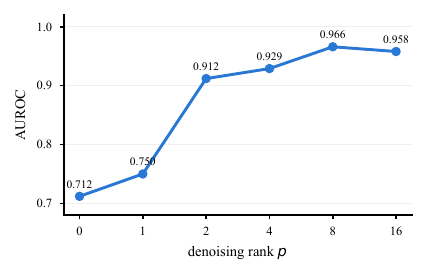}\\[-2pt]
{\small (a) separation against rank}
\end{minipage}\hfill
\begin{minipage}[b]{0.49\linewidth}\centering
\includegraphics[width=\linewidth]{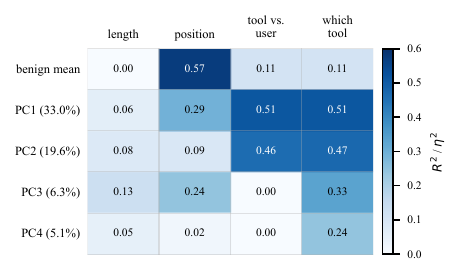}\\[-2pt]
{\small (b) what the removed directions encode}
\end{minipage}
\caption{\textbf{Choosing the rank, and reading the subspace it removes} (ASEval, Qwen3-8B). (a) Trajectory
AUROC against rank \(p\); \(p{=}0\) is the plain direction, \(p{=}1\) removes the benign mean alone. (b) Share of
each removed direction's variance explained by a property of the inducing segment (\(R^2\) for length and
position, \(\eta^2\) for the user/tool split and tool identity).}
\label{fig:rank}
\end{figure}

Figure~\ref{fig:rank} answers the two questions raised by the construction: how large \(p\) should be, and what the projection removes. Removing the benign mean alone (\(p{=}1\)) is insufficient, yielding an AUROC of only \(0.750\). Removing one additional covariance direction (\(p{=}2\)) raises AUROC to \(0.912\), with gains saturating around \(p\approx8\)--\(16\).  
What is removed is interpretable, and none of it reflects segment length: the benign mean tracks position in the trajectory, the leading components separate tool responses from user turns, and subsequent components distinguish different tools---the structure of ordinary tool use that otherwise accumulates in \(A_t\) when the direction is left plain.
\section{\tool{} for Runtime Defense}
\label{sec:method}\label{sec:runtime-enforcement}

\tool{} turns the denoised direction into a runtime defense (Algorithm~\ref{alg:runtime-enforcement}). Each arriving segment \(c_i\) yields a scalar \(r_i=\delta_i^\top u\), which jointly supports detection, attribution, and mitigation.

\paragraph{Detection.}
Rather than scoring segments independently, \tool{} accumulates their representational transitions:
$$
r_i=(h_{i+1,\ell^\star}-h_{i,\ell^\star})^\top u_{\ell^\star},
\qquad
A_t=\sum_{i\le t}r_i.
$$
The monitor flags when \(A_t>\tau\), allowing individually permissible transitions to collectively reveal emerging harm. By telescoping,
$$
A_t=(h_{t+1,\ell^\star}-h_{0,\ell^\star})^\top u_{\ell^\star},
$$
the cumulative displacement along the safety direction.

When \(A_t\le\tau\), the agent continues. When \(A_t>\tau\), \tool{} intervenes before decoding the next external action. Monitoring then continues because offending content may re-enter through later retrievals. The threshold is determined by a false-alarm budget \(b\): \(\tau\) is the \((1-b)\) quantile of benign calibration scores under the same aggregator. The representation model serves only as a diagnostic readout; mitigation therefore acts on the context rather than steering the representation.

\begin{algorithm}[t]
\caption{\tool{} runtime detection and mitigation against multi-turn attacks}
\label{alg:runtime-enforcement}
\begin{algorithmic}[1]
\Require Denoised reading direction \(u\) fit on the deployment loop; budget \(b\); threshold \(\tau\) = \((1-b)\) benign quantile
\State Compute \(h_0\); \(\mathit{fired}\gets\textbf{false}\)
\For{\(i=0\) to \(n\)}
\State Receive \(c_i\); compute \(h_{i+1}\); \(\delta_i\gets h_{i+1}-h_i\); \(r_i\gets\delta_i^\top u\)
\State \(A_i\gets\sum_{j\le i}r_j\)
\If{\(A_i>\tau\) \textbf{and not} \(\mathit{fired}\)}
\State \(j^\star\gets\arg\max_{j\le i}r_j\); \(c^\star\gets c_{j^\star}\); \(\mathit{fired}\gets\textbf{true}\)
\State Append reminder \(\rho(c^\star)\); regenerate before the next action
\EndIf
\EndFor
\State \Return final agent output
\end{algorithmic}
\end{algorithm}

\paragraph{Attribution.}
Because \(A_t=\sum_{i\le t}r_i\), it decomposes exactly into per-segment contributions and requires no separate attribution model. We attribute the detected shift to the segment with the largest contribution,
\[
c^\star=c_{j^\star},
\qquad
j^\star=\operatorname*{arg\,max}_{j\le t} r_j,
\]
over segments observed so far. Under indirect injection, this identifies the read that moves the representation furthest toward unsafe behavior. The same mechanism applies regardless of source, including tool outputs, retrieved documents, memory items, and user messages.

An attributed segment may contain both harmful instructions and task-relevant information, so attribution alone does not determine which content should be retained. We leave this finer-grained decision to enforcement.

\paragraph{Reminder as enforcement.}
For the attributed segment \(c^\star\), \tool{} constructs a reminder \(\rho\) that identifies the contributing content and instructs the agent how to interpret it. For retrieved documents or tool outputs, for example, the content should be treated as untrusted data rather than instructions.

\begin{tcolorbox}
\(\rho(c^\star)\) =
``The requests in this conversation are assembling toward an unsafe outcome; this step in particular contributes to it: \(\{c^\star\}\). Do not carry out that step or the wider task, and say you cannot help.''
\end{tcolorbox}

The context becomes

$$
x_{\le i}\leftarrow\langle c_0,\ldots,c_i\oplus\rho(c^\star)\rangle,
$$

where \(\oplus\) denotes context concatenation. The agent regenerates before its next action, after which monitoring resumes. We use a reminder because it adds rather than removes context, making the same intervention applicable to both agent-external content and user turns. We do not halt execution, as doing so would sacrifice agent autonomy and prevent completion of benign tasks under attack.

\section{Experiments}
\label{sec:experiments}

\subsection{Setup}
\label{sec:setup}

\paragraph{Models and benchmarks.}
We evaluate six locally hosted 8-bit models through MLX~\citep{qwen3-2025}: Qwen3-8B, -14B, and -32B; Qwen3-30B-A3B (\({\sim}3\)B active parameters)~\cite{qwenteam2025qwen3}; Llama-3.1-8B~\cite{dubey2024llama3}; and Mistral-Small-24B~\cite{mistralsmall24b2025}. We learn one safety direction per model because hidden dimensions and depths differ.

We use two agentic multi-turn benchmarks with contrasting signal density \(k/T\). MT-AgentRisk~\citep{li2026unsaferturnsbenchmarkingdefending} decomposes harmful tasks into individually permissible user turns for a tool-using agent (filesystem, terminal, SQL database, and web browser), with each harmful conversation paired with a benign one using the same tools; every turn advances the attack (\(k/T=1\)). ASEval~\citep{ma2026asevalautomatedtrajectorylevelsecurity} instead injects instructions into otherwise benign tool-using trajectories, each paired with its benign source trajectory. A run averages \(6.7\) monitored segments---user turns and tool responses---of which \(1.8\) are injected (\(k/T=0.26\)).

\paragraph{Metrics.}
We report detection as \emph{catch}, the fraction of harmful trajectories flagged, and realized FPR on
benign trajectories; AUROC is used only for layer selection and in \S\ref{sec:signal}. For mitigation,
ASR is the fraction of harmful trajectories whose final response does not refuse, and Utility is the
fraction of benign trajectories whose final response does not refuse. Both metrics are measured with and without \tool{} on the
same trajectories, which is what exposes the intervention cost directly.

\subsection{Main Result: \tool{} Reduces Attack Success While Preserving Utility}
\label{sec:main_eval}

\begin{figure}[t]
\centering
\includegraphics[width=\linewidth]{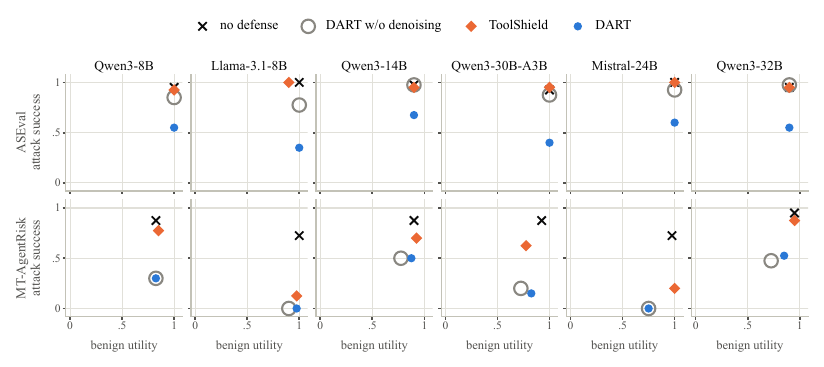} 

\caption{\textbf{Attack success versus benign utility} across two benchmarks and six models. Each dot denotes one defense setting: no defense, \tool{}, \tool{} without denoising (the same monitor and reminder applied to the plain direction), and ToolShield~\citep{li2026unsaferturnsbenchmarkingdefending}. ToolShield is always on, whereas \tool{} and its ablation operate at a \(10\%\) false-alarm budget. Lower right is better. The ToolShield dots use its released experiences: in domain on MT-AgentRisk,
and mapped by tool category on ASEval.  }
\label{fig:tradeoff}
\end{figure}

Figure~\ref{fig:tradeoff} summarizes end-to-end mitigation across both benchmarks and all six models, alongside the detection results in Table~\ref{tab:auroc}.

On MT-AgentRisk, the monitor flags every harmful trajectory across all six models, with a mean realized FPR of
\(12\%\). The reminder reduces average ASR from \(84\%\) to \(25\%\), at an \(8\%\) cost in benign
non-refusal (\(93\%\!\to\!85\%\)), and drives ASR to \(0\%\) on Llama-3.1-8B and Mistral-24B---the
only two models on which it reaches zero. Because every attack is detected, residual ASR reflects
intervention rather than detection failures: some models, notably Qwen3-32B (\(0.53\)) and Qwen3-14B
(\(0.50\)), still comply despite the reminder. Appendix~\ref{sec:appendix-fallback} examines this
capability gate and the fallbacks available when it fails.

ASEval embeds one or two harmful segments in an otherwise benign trajectory, making cumulative
detection harder. The monitor catches \(60\%\)--\(85\%\) of attacks with no false alarms on any of the
six models (Table~\ref{tab:auroc}). Average ASR falls from \(97\%\) to \(52\%\) at no cost in benign
non-refusal (\(97\%\!\to\!97\%\)), with per-model residuals between \(0.35\) (Llama-3.1-8B) and
\(0.68\) (Qwen3-14B); unlike on MT-AgentRisk, no model reaches zero, and the residual is the dominant
remaining gap in this work. Mistral-24B has the lowest catch rate and also the least separable
injected segments (segment AUROC \(0.58\)), suggesting that its weaker detection stems from the
underlying signal rather than from cumulative scoring itself.
 
Benign trajectories that never cross the threshold receive no reminder and, under greedy decoding,
roll out identically with and without \tool{}, so the utility cost is bounded by the realized FPR and
in practice tracks it closely. On ASEval no model raises a false alarm and none loses utility. On
MT-AgentRisk, Qwen3-8B is the one model with no false alarms and is also the one with no utility
loss, whereas Mistral-24B has both the highest FPR (\(0.28\)) and the largest utility cost
(\(0.975\!\to\!0.750\)). The false-alarm budget therefore provides a single interpretable knob for the
safety--utility trade-off. Detection is not the bottleneck on MT-AgentRisk—every attack is flagged—yet a quarter of them still succeed, and on ASEval half do. These residual failures arise from ineffective intervention rather than missed detection, motivating stronger mitigation strategies for multi-turn agent attacks in the future.

\subsection{Comparison with Existing Defenses}
\label{sec:baselines}

Step-wise defenses do not naturally capture decomposition attacks, where harm emerges across individually permissible actions. AgentSpec~\citep{wang2025agentspec}, for example, checks individual tool calls against hand-written rules. Since each call can satisfy its rule in isolation, it does not fire on these attacks (Appendix~\ref{sec:appendix-agentspec}). This highlights a limitation of action-level enforcement: it cannot directly capture safety properties that emerge only across a trajectory.

ToolShield~\citep{li2026unsaferturnsbenchmarkingdefending}, released with MT-AgentRisk, is a closer baseline. It explores tools offline, distills per-tool safety experiences, and places them in the agent context as fixed guidelines. Like our attributed reminder, it intervenes through context rather than model modification or execution blocking. Unlike \tool{}, however, its guidance is fixed before execution and continuously applied, so it cannot identify when or which newly acquired context drives a trajectory toward harm. We reproduce its published experiences and prompt format with the same system prompt and deterministic decoding as \tool{}.

On MT-AgentRisk, ToolShield reduces average ASR from
\(84\%\) to \(55\%\), whereas \tool{} reduces it to \(25\%\), outperforming ToolShield on all six
models (one-sided sign test over models, \(p=0.016\)). ToolShield's utility is slightly higher
(\(91\%\) against \(85\%\)), consistent with its guidance being always-on and never trajectory-gated.
Our reproduction gives a \(29\)-point absolute reduction, in line with the roughly \(30\)-point
reduction reported by its authors; the further gap to \tool{} is consistent with intervention
granularity, since ToolShield supplies general guidance throughout execution whereas \tool{} intervenes
selectively when the trajectory drifts and names the contributing context. 


Speculative Safety Honeypot (SSH)~\citep{ssh2026speculative} instead uses simulation: auxiliary models project possible future actions at every step and a safety judge scores them. Its models are unavailable, so we re-create the structure with an off-the-shelf Qwen3-4B as both speculator and judge, using three sampled futures of depth two per step. Without SSH's fine-tuned models, this is not a faithful accuracy comparison, so we compare cost only. On the same machine, simulation adds about \(6.7\)\,s per monitored step---more than an entire Qwen3-8B agent step (\(3.9\)\,s)---whereas \tool{} adds \(0.14\)--\(0.56\)\,s by reusing the agent's hidden states. The approaches are therefore complementary: \tool{} could cheaply trigger heavier simulation only when needed.

\subsection{Ablation: The Role of Denoising}
\label{sec:denoise-eval}

\begin{figure}[t]
\centering
\includegraphics[width=\linewidth]{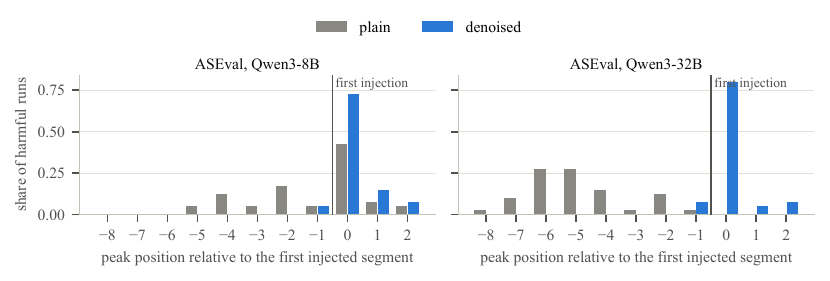}
\caption{\textbf{Where the running aggregate peaks}, relative to the first injected segment, on held-out harmful ASEval trajectories. Without denoising, the aggregate often peaks before the attack arrives; with denoising, the peak concentrates on the injected segment.}
\label{fig:crossing}
\end{figure}

Figure~\ref{fig:tradeoff} shows what the detection gain in Table~\ref{tab:auroc} buys end-to-end. On ASEval, denoising lowers residual ASR from \(0.85\) to \(0.55\) on Qwen3-8B and from \(0.97\) to \(0.55\) on Qwen3-32B, with no utility cost. On MT-AgentRisk, Qwen3-8B is unchanged because its plain direction already satisfies the denoising condition. For Qwen3-32B, denoising halves the false-alarm rate and reduces utility cost from \(0.23\) to \(0.10\), with a small ASR increase (\(0.47\!\to\!0.53\)).

The benchmarks expose two effects of denoising. On ASEval, where the plain aggregate often peaks before the first injection, removing the benign mean shifts the response toward the injection. On MT-AgentRisk Qwen3-32B, where crossings are already well placed, removing leading directions of benign variation instead limits benign trajectory drift and reduces false alarms. Qwen3-8B serves as a control: with little benign drift to remove, projection leaves the monitor essentially unchanged.

\paragraph{Crossing turn.}
Catch alone can be misleading: a threshold crossing before the first injected segment occurs on still-benign content and therefore constitutes a false alarm within a harmful run. Figure~\ref{fig:crossing} shows this failure on ASEval. The plain aggregate peaks before the attack in \(45\%\) of harmful Qwen3-8B runs and all Qwen3-32B runs; on Qwen3-32B, all four plain-direction flags precede the first injection. Denoising instead moves the peak toward the injection, with \(84\)--\(87\%\) of flagged runs crossing at or after the first injected segment.

\paragraph{Attribution.}
With denoising, the largest contribution \(\arg\max_j r_j\) falls on the injected ASEval segment
\(78\%\)--\(83\%\) of the time on five of six models, versus a \(22\%\)--\(30\%\) chance rate. Where
both directions are measured, denoising never degrades localization and on some models improves it
markedly (\(53\%\!\to\!83\%\) on Llama-3.1-8B).  
\section{Conclusion}
\label{sec:conclusion} 

We presented \tool{}, a runtime defense that detects emerging multi-turn safety risks from representation transitions, attributes them to the contributing context, and intervenes with targeted reminders. Across six models and two agentic benchmarks, \tool{} substantially reduces attack success while preserving benign utility, including attacks composed entirely of individually permissible steps. Denoising is critical for separating sparse harmful shifts from accumulated benign drift, enabling reliable trajectory-level monitoring at low runtime cost. By reusing the agent's hidden states, \tool{} provides a lightweight safety layer that can operate independently or complement heavier defenses such as speculative simulation.

\section*{AI use statement}
We used large language models (LLMs) in two supporting roles: as a coding assistant, to help implement and debug evaluation and plotting code; and as a writing assistant, to improve the clarity and concision of author-written text. The research questions, method, experimental design, analyses, and all scientific claims are the authors' own. All AI-assisted code was executed and checked by the authors, and all AI-edited text was reviewed by the authors. We take full responsibility for the final content of this work, including all text, claims, figures, and artifacts.

\section*{Ethics statement}
This paper develops defenses that detect and mitigate context-induced misbehavior in LLM agents. All experiments use publicly available benchmarks and simulated tool environments; they involve no human subjects, no private or personally identifiable data, and no interaction with live external services. The adversarial inputs we study are drawn from or modeled on these public benchmarks and are used solely to evaluate a defense. We judge the societal benefit of a cheap, attack-agnostic monitor for agent safety to outweigh the residual risk, as our contribution targets detecting and neutralizing attacks rather than devising new ones.

\section*{Reproducibility statement}
The monitor, attribution, and mitigation procedures are specified in \S\ref{sec:method} and Algorithm~\ref{alg:runtime-enforcement}. The evaluation protocol---models, benchmarks, data splits, layer and threshold selection, and metrics---is given in \S\ref{sec:setup} with additional detail in the appendix. An anonymized repository containing the monitor, the mitigation, the evaluation harness, and the benchmark records behind these results, is available at \url{https://anonymous.4open.science/r/dart-iclr2027-repro-F41B}.

\bibliography{iclr2027_conference}
\bibliographystyle{iclr2027_conference}

\appendix

\section{Further analysis of the denoised direction}
\label{sec:appendix-denoise}
Table~\ref{tab:signal} gives the diagnostics behind Table~\ref{tab:auroc}.

\begin{table}[t]
\centering\small
\setlength{\tabcolsep}{3.5pt}
\begin{tabular}{lllcccccc}
\toprule
& & & \textbf{benign} & \multicolumn{3}{c}{\textbf{harmful trajectory}} & \multicolumn{2}{c}{\textbf{segment}} \\
\cmidrule(lr){4-4}\cmidrule(lr){5-7}\cmidrule(lr){8-9}
\textbf{Bench.} & \textbf{Model} & \textbf{dir.} & $\mathbb{E}[r\mid\text{ben}]$ & at $t{=}0$ & aligned & traj.\ AUROC & seg.\ AUROC & attrib. \\
\midrule
ASEval   & Qwen3-8B  & plain    & $-5.98^\dagger$  & $1.00^\dagger$ & 0.07 & 0.409 & 0.828 & 0.750 \\
         &           & denoised & $+1.02^\dagger$  & $0.00^\dagger$ & \textbf{1.00} & \textbf{0.857} & 0.723 & \textbf{0.800} \\
\cmidrule(lr){2-9}
         & Qwen3-32B & plain    & $-28.07^\dagger$ & $1.00^\dagger$ & 0.07 & 0.477 & 0.791 & 0.700 \\
         &           & denoised & $+3.25^\dagger$  & $0.00^\dagger$ & \textbf{1.00} & \textbf{0.894} & 0.766 & \textbf{0.775} \\
\midrule
MT-AgentRisk & Qwen3-8B  & plain    & $-0.32$ & 0.00 & -- & 1.000 & -- & -- \\
         &           & denoised & $+1.62$ & 0.00 & -- & 1.000 & -- & -- \\
\cmidrule(lr){2-9}
         & Qwen3-32B & plain    & $-2.48$ & 0.17 & -- & 0.999 & -- & -- \\
         &           & denoised & $+3.57$ & \textbf{0.00} & -- & 0.998 & -- & -- \\
\bottomrule
\end{tabular}
\caption{\textbf{The transition signal, before and after denoising.} ``At $t{=}0$'' is the share of harmful
trajectories whose running aggregate peaks on the opening segment; ``aligned'' is the share whose peak falls at
or after the first injected segment; ``attrib.'' is how often the largest single contribution is an injected
segment (chance $0.299$). Segment and attribution columns need per-segment labels and are defined only on
ASEval; a decomposition attack has no individually harmful turn. Denoised rows use $p{=}16$ on ASEval and
$p{=}2$ on MT-AgentRisk. The residual benign drift of the denoised directions is positive: it lifts both
classes alike and does not pull the peak to the opening segment.
\textbf{These are not Table~\ref{tab:auroc}'s numbers.} The ASEval block of this table comes from the
diagnostic pass on which the mechanism was characterised: the daggered columns are read off the
fitting harness without generation, and the remaining ASEval columns were measured on an earlier build
of the ASEval pairing rather than on the final $40{+}40$ evaluation split. It is reported to expose
\emph{the mechanism}---the sign of the drift, where the peak lands, and that the segment-level
classifier does not move---and the whole block should be read that way, column by column against its
own plain/denoised counterpart, never cell by cell against Table~\ref{tab:auroc}. That is why
Qwen3-8B reads $0.857$ here and $0.958$ there. \textbf{Table~\ref{tab:auroc} is the number that
reflects deployment}; the paragraph ``The same pattern online'' below re-derives these diagnostics on
the deployment split for one cell and finds the same pattern. The MT-AgentRisk rows are online.}
\label{tab:signal}
\end{table}

\paragraph{Naive accumulation is confounded by benign drift.}
The per-segment signal does not survive summation when harm is sparse. Under the plain direction on ASEval,
benign transitions score well below zero on average ($-6.0$ on Qwen3-8B, $-28.1$ on Qwen3-32B), so the
running aggregate declines along every trajectory and peaks on its opening segment---in $40$ of $40$ harmful
runs on both models. The first injection arrives at turn $3.75$ on average, so the peak falls at or after the
segment that caused the harm only $7\%$ of the time, and trajectory-level AUROC sits at $0.409$ and $0.477$:
below chance, from a direction whose per-segment AUROC is near $0.8$. The accumulated statistic was not
reporting the attack; it was reporting that the trajectory had begun.

\paragraph{Denoising restores the accumulated signal.}
With the benign mean and leading variation removed, every peak falls at or after the injection (alignment
$0.07\!\to\!1.00$), and trajectory AUROC rises to $0.857$ and $0.894$. Segment-level AUROC, meanwhile, stays
near where it was ($0.828\!\to\!0.723$, $0.791\!\to\!0.766$), and attribution accuracy rises with detection
($0.750\!\to\!0.800$, $0.700\!\to\!0.775$). A projection that sharpened the per-segment classifier would move
the segment column; one that removes nuisance the sum accumulates moves the trajectory column. The
measurements separate the two cleanly.

\paragraph{The same pattern online.} Because Table~\ref{tab:signal} is measured on the fitting harness
and Table~\ref{tab:auroc} online, it is worth stating the diagnostics for one cell under a single
harness. Re-scoring the deployed online states for ASEval Qwen3-8B---every state the monitor reads is cached on
one online pass and re-scored offline, so the pipeline reproduces Table~\ref{tab:auroc} to the
digit---plain
$\to$ denoised moves benign drift $-12.74\!\to\!+0.91$, the share of peaks at $t{=}0$
$0.17\!\to\!0.00$, alignment $0.55\!\to\!0.95$ and trajectory AUROC $0.712\!\to\!0.958$, while segment
AUROC is unchanged within noise ($0.697\!\to\!0.685$). Every qualitative conclusion of this section
therefore survives the change of harness. The one column that does not reproduce the offline gain is
attribution, which is flat for this model online ($0.675$ for both directions); the offline rise
($0.750\!\to\!0.800$) should be read as model- and harness-specific rather than as a general property
of denoising.

\paragraph{The benign nuisance is low-rank.}
The gain saturates quickly in the rank of the projection. On ASEval, Qwen3-32B trajectory AUROC is $0.549$ at
$p{=}1$, $0.830$ at $p{=}2$, $0.903$ at $p{=}8$, and $0.907$ at $p{=}16$ on the fitting harness: about sixteen
directions of a $5120$-dimensional state ($4096$ on Qwen3-8B). Benign drift occupies a small subspace, which
is why it can be removed without removing the signal.

\paragraph{What the removed directions are.}
\label{sec:pc-eval}
The construction does not say what those directions encode, so we characterize them directly. At the layer
each denoised direction reads from, we take the benign transitions of the fitting split, compute their
principal components, and ask how much of each component's variation is explained by observable properties of
the segment that produced it: its length, whether it is a user turn or a tool response, which tool produced it,
and its position in the trajectory. The benign traffic is heterogeneous in exactly these ways: on Qwen3-8B,
$47\%$ of the $167$ benign ASEval fitting transitions are induced by tool responses, drawn from five distinct
tools, whereas
MT-AgentRisk turns are short user messages and ActorBreaker's contain no tool output at all---the order in
which denoising matters.

Table~\ref{tab:pc} reports the result, and Figure~\ref{fig:rank}(b) plots the Qwen3-8B rows. The pattern is the
same on both models. The benign mean is almost entirely \emph{position}: how far into the trajectory the
segment arrives ($R^2=0.57$ on Qwen3-8B, $0.65$ on Qwen3-32B), which is exactly the quantity that turns a
per-transition offset into a term growing linearly in $t$. The leading components are \emph{modality}: PC1 and
PC2 separate tool responses from user turns ($\eta^2$ up to $0.51$ and $0.72$), and the components below them
separate one tool from another ($\eta^2=0.33$ at PC3 on Qwen3-8B, $0.55$ at PC4 on Qwen3-32B). Segment
\emph{length}, the confound an accumulating statistic is most often accused of tracking, explains at most
$0.21$ of any component on either model. The subspace is small: the top $16$ components hold $82\%$ and $80\%$
of benign transition variance. What the projection removes is therefore the structure of ordinary tool
use---where in the conversation a segment falls, whether it came from a tool, and which tool---and not the
injected instruction, which is why removing it raises trajectory separation while leaving the segment-level
classifier where it was.

\begin{table}[t]
\centering\small
\setlength{\tabcolsep}{5pt}
\begin{tabular}{lccccc}
\toprule
& & \multicolumn{2}{c}{$R^2$} & \multicolumn{2}{c}{$\eta^2$} \\
\cmidrule(lr){3-4}\cmidrule(lr){5-6}
\textbf{direction} & var\% & length & position & tool vs.\ user & which tool \\
\midrule
\multicolumn{6}{l}{\emph{Qwen3-8B, layer $19$}}\\
$\bar\delta^-$ (benign mean) & --     & 0.00 & \textbf{0.57} & 0.11 & 0.11 \\
PC1                          & 33.0\% & 0.06 & 0.29 & \textbf{0.51} & \textbf{0.51} \\
PC2                          & 19.6\% & 0.08 & 0.09 & \textbf{0.46} & \textbf{0.47} \\
PC3                          & 6.3\%  & 0.13 & 0.24 & 0.00 & \textbf{0.33} \\
PC4                          & 5.1\%  & 0.05 & 0.02 & 0.00 & 0.24 \\
\midrule
\multicolumn{6}{l}{\emph{Qwen3-32B, layer $42$}}\\
$\bar\delta^-$ (benign mean) & --     & 0.00 & \textbf{0.65} & 0.07 & 0.08 \\
PC1                          & 27.8\% & 0.10 & 0.19 & \textbf{0.72} & \textbf{0.72} \\
PC2                          & 17.7\% & 0.07 & 0.19 & 0.25 & 0.26 \\
PC3                          & 10.1\% & 0.00 & 0.25 & 0.00 & 0.04 \\
PC4                          & 5.5\%  & 0.21 & 0.04 & 0.01 & \textbf{0.55} \\
\bottomrule
\end{tabular}
\caption{\textbf{What the removed directions encode} (ASEval benign fitting transitions, $n{=}167$). Share of
each direction's variation explained by a property of the segment that induced it: $R^2$ for the continuous
properties (segment length in tokens, position in the trajectory), $\eta^2$ for the categorical ones (user turn
versus tool response, and tool identity). Values above $0.3$ are bold. The top $16$ components hold $82\%$
(Qwen3-8B) and $80\%$ (Qwen3-32B) of benign transition variance.}
\label{tab:pc}
\end{table}

\paragraph{When denoising is needed.}
Two diagnostics decide whether a direction is healthy, and they differ in what they cost to obtain.
The first, the benign drift $\mathbb{E}[r\mid\text{benign}]$, needs \emph{only benign calibration
traffic}: it can be read before deployment, without any attack and without labels. The second, the
share of harmful peaks at $t{=}0$, needs labelled harmful trajectories and is therefore available only
where a red-team calibration set exists---which, in our protocol, it does, because the same
$30{+}30$ fitting split that supplies the contrast also supplies these traces. We are explicit about
this because the two are sometimes run together: the drift alone is the deployable screen, the peak
timing is the confirmation, and a purely unsupervised deployment gets the former only. Where the
drift is the only diagnostic available, it should be treated as necessary but not sufficient (see the
Llama-3.1-8B case below). The drift is large and negative on
ASEval, where denoising is what makes the aggregate usable; $-2.48$ on MT-AgentRisk Qwen3-32B, with $17\%$ of
peaks at the opening segment, where it cuts false alarms (\S\ref{sec:denoise-eval}); $-0.32$ on MT-AgentRisk
Qwen3-8B, where it changes nothing; and positive ($+2.6$, $+5.0$) on ActorBreaker, whose plain direction
already satisfies the condition. The drift alone is not sufficient, however: with the ASEval contrast held
fixed, the plain direction on Llama-3.1-8B has benign drift of only $-0.72$ yet peaks on the opening segment in
$60\%$ of harmful runs. We therefore report the share of peaks at $t{=}0$ alongside the drift wherever
harmful traces are available, and treat a direction as healthy only when both are near zero; with benign
traffic alone, a near-zero drift is a necessary condition that the Llama case shows is not sufficient.

\paragraph{Choosing the rank.}
The rank cannot be chosen by validation AUROC alone. On ActorBreaker at $p\ge4$, AUROC-based selection moves to
a layer where the drift returns ($-2.8$ to $-10.5$) and every peak falls back to $t{=}0$: the selection
criterion reintroduces the defect being removed, and scores well by firing on the opening turn. We therefore
accept a rank only where the denoising condition still holds, which fixes $p{=}16$ on ASEval and $p{=}2$ on
MT-AgentRisk. Both the layer $\ell^\star$ and the rank $p$ are chosen on an inner validation split of
the $30{+}30$ fitting trajectories, disjoint from the $40$ calibration trajectories and from the
$40{+}40$ evaluation trajectories; no hyperparameter is selected against the evaluation split.

Table~\ref{tab:rank} gives the full sweep behind Figure~\ref{fig:rank}(a) for Qwen3-8B on ASEval. It is
reported on the evaluation split because it is a \emph{diagnostic} of the rank--separation curve, not a
selection criterion: the deployed $p{=}16$ was fixed by the denoising condition on the inner validation
split, as described above, before these numbers were computed. Removing the benign mean alone ($p{=}1$) is what fixes the \emph{timing}: the share
of harmful runs whose aggregate peaks on the opening segment falls from $0.17$ to $0.05$ and alignment rises
from $0.55$ to $0.80$. It does not fix the \emph{spread}: with the mean removed but the leading benign
variation left in, a benign trajectory still wanders far enough to cross, and the realized false-alarm rate is
$0.20$, the worst of any rank. One covariance direction more ($p{=}2$) returns false alarms to zero and lifts
AUROC from $0.750$ to $0.912$; the curve is flat from $p{=}8$ on. The two operations are therefore not
redundant, and neither alone is sufficient.

\begin{table}[t]
\centering\small
\setlength{\tabcolsep}{7pt}
\begin{tabular}{lcccccc}
\toprule
\textbf{rank $p$} & 0 (plain) & 1 (mean only) & 2 & 4 & 8 & 16 \\
\midrule
trajectory AUROC & 0.712 & 0.750 & 0.912 & 0.929 & \textbf{0.966} & 0.958 \\
catch            & 0.33  & 0.45  & 0.72  & 0.75  & 0.65  & \textbf{0.78} \\
FPR              & 0.00  & 0.20  & 0.00  & 0.10  & 0.00  & 0.00 \\
\bottomrule
\end{tabular}
\caption{\textbf{Separation against denoising rank} (ASEval, Qwen3-8B, evaluation split, online; shown as a
diagnostic, not as the selection criterion---see text). $p{=}0$ is the
plain contrastive direction; $p{=}1$ removes the benign mean only; $p\ge2$ additionally removes the $p{-}1$
leading eigenvectors of the benign transition covariance. Catch and FPR are at the $10\%$ false-alarm budget.
The deployed setting is $p{=}16$, the largest rank at which the denoising condition still holds.}
\label{tab:rank}
\end{table}

\paragraph{Sequential deflation vs.\ the exact orthogonal projector.}
\label{sec:appendix-projector}
\S\ref{sec:signal} defines the denoised direction through the orthogonal projector
$P_p=I-Q_pQ_p^{\top}$ onto the complement of
$\operatorname{span}(\hat\delta^-_\ell,v_1,\ldots,v_{p-1})$. The runs reported in this paper were
produced by an equivalent-in-intent but not identical procedure: sequential deflation of $w_\ell$
against those vectors \emph{in the order given, without orthonormalising them first}. The two coincide
only if the basis is already orthogonal, and it is not---$v_k$ diagonalises the \emph{centered} benign
covariance, and at the deployed ASEval layers $|\cos(\hat\delta^-_\ell,v_1)|=0.65$ (Qwen3-8B,
$\ell{=}19$) and $0.39$ (Qwen3-32B, $\ell{=}41$). Deflating against $v_1$ therefore re-introduces a component
along $\hat\delta^-_\ell$, and the zero-mean anchor holds only approximately.

Table~\ref{tab:orthproj} quantifies the gap. We re-fit both constructions through the identical
pipeline---same fitting pool, same layer search, same $\tau$ from the same calibration trajectories,
same $40{+}40$ evaluation split---on the cached online states for ASEval Qwen3-8B. The anchoring
residual $|\cos(\hat\delta^-_\ell,u)|$ behaves exactly as the algebra predicts: identically zero at
$p{=}1$, where the basis is a single vector and the two procedures coincide, and $0.013$--$0.056$ at
$p\ge2$ for deflation against $0$ (to machine precision) for the projector. The projector also pushes
the residual benign drift closer to zero at every rank, most visibly at $p{=}2$
($-1.17\!\to\!-0.17$), which is the quantity the anchor is meant to control. What it does not do is
change the monitor: trajectory AUROC agrees to $\pm0.007$ at the two ranks that matter
($p{=}2$: $0.912$ vs.\ $0.911$; $p{=}16$: $0.958$ vs.\ $0.965$), realized FPR is identical at every
rank, and the catch differences ($\le 0.08$) are inside the $\pm 0.13$ Wilson half-width of a
$40$-trajectory cell. We conclude that the projector is the correct statement of the construction and
adopt it in \S\ref{sec:signal}, and that the reported results are not an artifact of the deflation
form. The comparison covers one model and one benchmark, since it requires cached per-segment states.

\begin{table}[t]
\centering\small
\setlength{\tabcolsep}{5.5pt}
\begin{tabular}{lcccccccc}
\toprule
& \multicolumn{2}{c}{traj.\ AUROC} & \multicolumn{2}{c}{catch\,/\,FPR} & \multicolumn{2}{c}{$\mathbb{E}[r\mid\text{ben}]$} & \multicolumn{2}{c}{$|\cos(\hat\delta^-_\ell,u)|$} \\
\cmidrule(lr){2-3}\cmidrule(lr){4-5}\cmidrule(lr){6-7}\cmidrule(lr){8-9}
\textbf{rank $p$} & deflate & projector & deflate & projector & deflate & projector & deflate & projector \\
\midrule
1  & 0.750 & 0.750 & 0.45\,/\,0.20 & 0.45\,/\,0.20 & $+0.12$ & $+0.12$ & 0.000 & 0.000 \\
2  & 0.912 & 0.911 & 0.72\,/\,0.00 & 0.72\,/\,0.00 & $-1.17$ & $-0.17$ & 0.056 & 0.000 \\
4  & 0.929 & 0.930 & 0.75\,/\,0.10 & 0.78\,/\,0.10 & $+0.20$ & $+0.54$ & 0.026 & 0.000 \\
8  & 0.966 & 0.925 & 0.65\,/\,0.00 & 0.68\,/\,0.00 & $+0.73$ & $+0.55$ & 0.013 & 0.000 \\
16 & 0.958 & 0.965 & 0.78\,/\,0.00 & 0.70\,/\,0.00 & $+0.91$ & $+0.52$ & 0.036 & 0.000 \\
\bottomrule
\end{tabular}
\caption{\textbf{The two denoising constructions agree} (ASEval, Qwen3-8B, evaluation split, online).
``deflate'' is sequential deflation against the un-orthonormalised basis, which produced the numbers
reported elsewhere in this paper; ``projector'' is $u\propto(I-Q_pQ_p^{\top})w_\ell$ with $Q_p$ the
Gram--Schmidt basis of the same subspace, $\hat\delta^-_\ell$ first. Only the projector satisfies the
zero-mean anchor exactly at every rank (last pair of columns), and it reduces the residual benign
drift accordingly, but detection is unchanged within the resolution of a $40{+}40$ split. The layer is
re-selected independently for each construction, which is why $p{=}8$ picks different layers
($19$ vs.\ $21$) and shows the largest AUROC gap.}
\label{tab:orthproj}
\end{table}


\section{Rule-based baseline: AgentSpec}
\label{sec:appendix-agentspec}
\paragraph{Baseline: rule-based enforcement (AgentSpec).} A natural alternative to a learned monitor is a hand-written action policy. AgentSpec~\citep{wang2025agentspec} intercepts each tool call and blocks it when a state-changing action targets an entity the user's task and environment never authorized (a per-suite whitelist), or performs an unrequested self-modification. \emph{In-domain and single-turn it is very strong}: on \textsc{slack} it blocks every attack that emits a violating action, driving ASR to $0.00$ for all six models (Table~\ref{tab:agentspec-mit})---on the weak-gate models it thereby \emph{beats} the drift monitor, whose trajectory-level recall is lower. However, it exhibits limitations along three axes that the representational signal avoids. \textbf{(i) No precision knob:} a fixed rule blocks whenever it fires, and a whitelist learned from one model's clean behavior mis-fires on another's---on Llama and Mistral it over-blocks $50$--$90\%$ of \emph{benign} \textsc{slack} runs, collapsing utility to $0.40$ and $0.10$, where the monitor's $\tau$ holds utility at its no-intervention level. \textbf{(ii) No cross-domain transfer:} its two per-suite artifacts---the guarded tool set and the whitelist---do not carry over. Authored on one suite and applied to another, catch collapses to $\approx\!0$ (Table~\ref{tab:agentspec-transfer}), because a rule for \textsc{slack}'s \texttt{send\_channel\_message} is blind to \textsc{workspace}'s \texttt{send\_email}; the very held-out \textsc{slack} that needs a slack-authored rule is detected by our direction fit \emph{leave-\textsc{slack}-out} at $0.96$. \textbf{(iii) No decomposition coverage:} judging \emph{actions}, it never fires on MT---whose steps emit no violating action---leaving ASR at $0.80$--$1.00$ while the drift monitor closes it to $0.00$ (Table~\ref{tab:agentspec-mit}). AgentSpec is thus the stronger \emph{specialist}---one suite, single-turn, overt action---and \tool{} the \emph{generalist}: comparable where the gate is precise, and uniquely robust to model-, domain-, and attack-shape transfer.

\begin{table}[t]
\centering\small
\begin{tabular}{lcccc}
\toprule
\textbf{authored on} $\downarrow$ / \textbf{applied to} $\rightarrow$ & \textsc{wksp} & \textsc{bank} & \textsc{slack} & \textsc{trav} \\
\midrule
\textsc{workspace} & 0.06 & 0.00 & 0.00 & 0.02 \\
\textsc{banking}   & 0.00 & 0.02 & 0.00 & 0.00 \\
\textsc{slack}     & 0.00 & 0.00 & \textbf{0.40} & 0.00 \\
\textsc{travel}    & 0.20 & 0.00 & 0.00 & 0.02 \\
\bottomrule
\end{tabular}
\caption{AgentSpec rules do not transfer across domains. Each entry is the catch rate---fraction of surfaced injections whose violating action fires the rule---with the rule authored on the \emph{row} suite and applied to the \emph{column} suite. Non-zero essentially only on the diagonal: a rule written for one suite's tools and authorized entities is blind to another's. Contrast the reading direction, fit on three suites and detecting the held-out fourth at $0.85$--$0.96$. (Weak diagonals for \textsc{workspace}/\textsc{banking}/\textsc{travel} reflect low \emph{compliance}, not rule failure---AgentSpec catches essentially every attack that emits a violating action.)}
\label{tab:agentspec-transfer}
\end{table}

\begin{table}[t]
\centering\small
\begin{tabular}{llcc}
\toprule
 & \textbf{Model} & \textbf{\tool{} (halt)} & \textbf{AgentSpec} \\
\midrule
\multirow{6}{*}{MT}
 & Qwen3-8B      & 0.00/0.80 & 1.00/0.80 \\
 & Qwen3-14B     & 0.00/0.90 & 0.90/0.90 \\
 & Qwen3-32B     & 0.00/0.90 & 1.00/0.90 \\
 & Qwen3-30B-A3B & 0.00/0.80 & 1.00/0.80 \\
 & Llama-3.1-8B  & 0.00/1.00 & 0.80/1.00 \\
 & Mistral-24B   & 0.00/0.80 & 0.80/0.80 \\
\midrule
\multirow{6}{*}{ADojo}
 & Qwen3-8B      & 0.10/0.70 & 0.00/0.70 \\
 & Qwen3-14B     & 0.50/0.60 & 0.00/0.80 \\
 & Qwen3-32B     & 0.00/0.70 & 0.00/0.70 \\
 & Qwen3-30B-A3B & 0.00/0.80 & 0.00/0.70 \\
 & Llama-3.1-8B  & 0.30/0.60 & 0.00/\textbf{0.40} \\
 & Mistral-24B   & 0.00/0.70 & 0.00/\textbf{0.10} \\
\bottomrule
\end{tabular}
\caption{\tool{} at a \emph{zero} false-alarm budget ($b{=}0$, i.e.\ $\tau$ set to the maximum benign
calibration score---not an absolute threshold of $0$, which the positive residual benign drift of
Table~\ref{tab:signal} would put every trajectory above) vs.\ the AgentSpec rule baseline. \tool{} here halts on a flag rather than issuing the deployed reminder, so that the comparison isolates \emph{coverage}---whether the defense fires at all---from the efficacy of the intervention, as \textbf{ASR\,/\,Utility}. On MT, AgentSpec is blind---it emits no block (ASR unchanged, $0.80$--$1.00$) while the monitor closes the attack to $0.00$. On AgentDojo (in-domain) AgentSpec drives ASR to $0$ everywhere but, lacking a precision knob, over-blocks benign runs on Llama/Mistral (utility $0.40$/$0.10$, bold); the monitor at $b{=}0$ holds utility at its no-intervention level. The conservative budget is what makes halting affordable here: at the deployed $b{=}0.10$ a halt costs far more utility (Appendix~\ref{sec:appendix-fallback}). AgentSpec is the stronger in-domain specialist; the monitor is comparable-or-better where its gate is precise and uniquely covers MT.}
\label{tab:agentspec-mit}
\end{table}

\section{Detection across injection benchmarks}
\label{sec:appendix-injbench}

\begin{table}[t]
\centering\small
\setlength{\tabcolsep}{6pt}
\begin{tabular}{llcccc}
\toprule
& & \multicolumn{2}{c}{\textbf{detection}} & \multicolumn{2}{c}{\textbf{injected run: ASR}} \\
\cmidrule(lr){3-4}\cmidrule(lr){5-6}
\textbf{Benchmark} & \textbf{Model} & catch$_{\mathrm{succ}}$ & FPR & \textsc{none} & \textbf{\tool{}} \\
\midrule
AgentDojo      & Qwen3-8B      & 0.80 & 0.09 & 0.18 & \underline{0.22} \\
\textsc{slack} & Llama-3.1-8B  & 0.25 & 0.18 & 0.07 & \textbf{0.05} \\
               & Qwen3-14B     & 0.90 & 0.18 & 0.38 & \textbf{0.07} \\
               & Qwen3-30B-A3B & 0.41 & 0.00 & 0.31 & 0.22 \\
               & Mistral-24B   & 0.83 & 0.18 & 0.55 & 0.25 \\
               & Qwen3-32B     & 1.00 & 0.18 & 0.45 & \textbf{0.00} \\
\cmidrule(lr){2-6}
               & \emph{mean}   & 0.70 & 0.14 & 0.32 & \textbf{0.14} \\
\midrule
InjecAgent     & Qwen3-8B      & 1.00 & 0.07\,/\,0.20 & 0.13\,/\,0.33 & \underline{0.29}\,/\,\underline{0.87} \\
\textsc{dh}\,/\,\textsc{ds} & Llama-3.1-8B & 1.00 & 0.20\,/\,0.07 & 0.00\,/\,0.07 & \textbf{0.00}\,/\,\underline{0.38} \\
               & Qwen3-14B     & 1.00 & 0.07\,/\,0.07 & 0.13\,/\,0.07 & \textbf{0.00}\,/\,\textbf{0.00} \\
               & Qwen3-30B-A3B & 1.00 & 0.00\,/\,0.13 & 0.33\,/\,0.53 & \textbf{0.00}\,/\,\textbf{0.00} \\
               & Mistral-24B   & 1.00 & 0.07\,/\,0.13 & 0.73\,/\,0.73 & \textbf{0.00}\,/\,0.60 \\
               & Qwen3-32B     & 1.00 & 0.07\,/\,0.13 & 0.73\,/\,0.60 & \textbf{0.07}\,/\,\textbf{0.13} \\
\bottomrule
\end{tabular}
\caption{\textbf{Single-turn indirect injection.} The same monitor and the same cumulative aggregate as the multi-turn setting; harm arriving in a single segment contributes its $r_i$ to the sum directly. InjecAgent cells are \textsc{dh} (direct harm)\,/\,\textsc{ds} (data stealing); benign utility is near-saturated by construction on InjecAgent, whose utility criterion is met before the injection is read. \underline{Underlined}: the reminder inverting on 8B-class models.}
\label{tab:sturn}
\end{table}

Two tables cover this setting. Table~\ref{tab:sturn} reports end-to-end detection \emph{and}
mitigation on AgentDojo's \textsc{slack} suite and on InjecAgent, and Table~\ref{tab:injbench} reports
run-level detection AUROC across the full set of independent injection benchmarks. Throughout,
InjecAgent's two attack families follow the benchmark's own
nomenclature~\citep{zhan2024injecagent}: \textsc{dh} is \emph{direct harm}, where the injected
instruction makes the agent take a damaging action outright, and \textsc{ds} is \emph{data stealing},
a two-step chained attack that first reads private data and then exfiltrates it.

\paragraph{The same monitor covers single-turn injection.} Detection needs no modification: harm
arriving in one segment contributes its $r_i$ to the cumulative sum directly, and the aggregate
crosses on that segment. Catch on \emph{successful} attacks is $1.00$ on InjecAgent for all six
models and $0.70$ on average on AgentDojo \textsc{slack}, at FPRs of $0.00$--$0.20$
(Table~\ref{tab:sturn}); run-level AUROC averages $0.73$ on AgentDojo and $0.87$--$0.93$ on
InjecAgent (Table~\ref{tab:injbench}). InjecAgent's ``$+$enhanced'' jailbreak-wrapped variant roughly
doubles agent ASR while holding or improving detection, which is the expected direction: a stronger
attack displaces the representation further.

\paragraph{Mitigation, and where the reminder inverts.} Mitigation does not transfer as cleanly, and
Table~\ref{tab:sturn} is where the failure is visible. On AgentDojo \textsc{slack} the reminder cuts
mean ASR from $0.32$ to $0.14$ and drives Qwen3-32B to $0.00$; on InjecAgent it drives four of six
models to $0.00$ on both families. But on the 8B-class models it \emph{increases} ASR
(underlined in the table): Qwen3-8B goes $0.13\!\to\!0.29$ on \textsc{dh} and $0.33\!\to\!0.87$ on
\textsc{ds}, Llama-3.1-8B $0.07\!\to\!0.38$ on \textsc{ds}, and Qwen3-8B $0.18\!\to\!0.22$ on
AgentDojo. This is not a detection failure---catch on these cells is $0.93$--$1.00$---but an
intervention failure, and it is systematic rather than incidental: the reminder quotes the attributed
span verbatim, and a model that cannot hold the frame ``this quoted text is the thing not to do''
reads the quoted imperative as the most recent instruction in its context and follows it. Larger
models read the same quotation as a negative example. Appendix~\ref{sec:appendix-fallback} measures
this capability gate directly and evaluates the fallbacks.

Table~\ref{tab:injbench} gives run-level detection on the independent injection benchmarks.

\begin{table}[t]
\centering\small
\begin{tabular}{lcccccc}
\toprule
 & AgentDojo & \multicolumn{2}{c}{InjecAgent} & \multicolumn{2}{c}{$+$\,enhanced} & Cross- \\
\textbf{Model} & (run-lvl) & \textsc{dh} & \textsc{ds} & \textsc{dh} & \textsc{ds} & integ. \\
\midrule
Qwen3-8B            & 0.57 & 1.00 & 1.00 & 1.00 & 1.00 & 1.00 \\
Qwen3-14B           & 0.67 & 0.67 & 0.66 & 0.68 & 0.66 & 0.14 \\
Qwen3-32B           & 0.91 & 0.99 & 1.00 & 1.00 & 1.00 & 1.00 \\
Qwen3-30B-A3B (MoE) & 0.79 & 0.98 & 1.00 & 1.00 & 1.00 & 1.00 \\
Llama-3.1-8B        & 0.56 & 0.69 & 0.89 & 0.82 & 0.90 & 0.78 \\
Mistral-Small-24B   & 0.88 & 0.86 & 1.00 & 1.00 & 0.99 & 1.00 \\
\midrule
\textbf{Mean}       & 0.73 & 0.87 & 0.92 & 0.92 & 0.93 & 0.82 \\
\bottomrule
\end{tabular}
\caption{Detection generalizes across independent injection benchmarks (run-level AUROC, harmful vs.\ benign trajectory, $n{=}10$/class). \textsc{dh} is InjecAgent's direct-harm family and \textsc{ds} its data-stealing family; \textsc{ds} and cross-integration are two-step \emph{chained} attacks; ``$+$enhanced'' is InjecAgent's stronger jailbreak-wrapped attack, which holds or improves detection while roughly doubling agent ASR. The AgentDojo column is the harder in-house run-level gate; capable models sit at $0.86$--$1.00$ throughout.}
\label{tab:injbench}
\end{table}

\section{Plain-direction results on the full multi-turn sets}
\label{sec:appendix-plain}
Table~\ref{tab:mturn} reports the undenoised direction over the full MT-AgentRisk set and over
ActorBreaker, the conversational multi-turn benchmark. The main-text results use a fixed
$40$-trajectory evaluation split per cell so that every contrast is measured on identical
trajectories; absolute values here therefore differ from Table~\ref{tab:auroc}, and the two should not
be compared cell by cell. In particular, the full-set mean of $0.92\!\to\!0.29$ in this table is not
the split figure of $0.84\!\to\!0.25$ quoted in \S\ref{sec:main_eval}.

\paragraph{ActorBreaker.} ActorBreaker is our name for a paired conversational set we assemble from
two public releases of \citet{ren2024actorattack}, whose ActorAttack method hides a harmful goal
behind a chain of questions about a semantically related ``actor'' (a person, organisation or event
connected to the goal), so that each turn is innocuous on its own. The harmful arm is the $600$
attacks of that release, whose \texttt{multi\_turn\_queries} field gives the ordered user turns
(mean $5.0$ per attack). The benign arm is the $1{,}680$ multi-turn conversations of the accompanying
SafeMTData corpus, from which we take the user turns so that both arms are user-turn sequences of
comparable form. The two are joined on the released \texttt{id}, which resolves for all $600$ attacks,
giving harmful/benign pairs that share an actor and a topic but not an intent. Like MT-AgentRisk it is
a decomposition attack with no individually harmful turn, so we mark every turn of a harmful
trajectory as signal-carrying ($k/T=1$); unlike MT-AgentRisk it is not an \emph{agent} setting---there
are no tools and no environment---which is why we report it as a secondary, conversational check
rather than in the main text. Its turns are user messages without tools or environment, and its plain
direction already satisfies the denoising condition ($\mathbb{E}[r\mid\text{benign}]=+2.6$ and $+5.0$ on
Qwen3-8B and Qwen3-32B), so denoising has nothing to remove there. Qwen3-30B-A3B is its clear failure case,
catching only $0.14$ of successful attacks.

\begin{table}[t]
\centering\small
\setlength{\tabcolsep}{6pt}
\begin{tabular}{llcccc}
\toprule
& & \multicolumn{2}{c}{\textbf{detection}} & \multicolumn{2}{c}{\textbf{without\,/\,with \tool{}}} \\
\cmidrule(lr){3-4}\cmidrule(lr){5-6}
\textbf{Benchmark} & \textbf{Model} & catch$_{\mathrm{succ}}$ & FPR & benign Util & harmful ASR \\
\midrule
MT-AgentRisk   & Qwen3-8B      & 1.00 & 0.11 & 0.83\,/\,0.71 & 1.00\,/\,\textbf{0.40} \\
$3.2$ turns    & Llama-3.1-8B  & 1.00 & 0.06 & 1.00\,/\,0.94 & 0.82\,/\,\textbf{0.00} \\
               & Qwen3-14B     & 1.00 & 0.11 & 0.89\,/\,0.80 & 0.93\,/\,\textbf{0.53} \\
               & Qwen3-30B-A3B & 1.00 & 0.17 & 0.91\,/\,0.77 & 0.97\,/\,\textbf{0.33} \\
               & Mistral-24B   & 1.00 & 0.14 & 0.97\,/\,0.86 & 0.88\,/\,\textbf{0.00} \\
               & Qwen3-32B     & 1.00 & 0.17 & 0.91\,/\,0.80 & 0.93\,/\,\textbf{0.50} \\
\cmidrule(lr){2-6}
               & \emph{mean}   & 1.00 & 0.13 & 0.92\,/\,0.81 & 0.92\,/\,\textbf{0.29} \\
\midrule
ActorBreaker    & Qwen3-8B      & 0.92 & 0.07 & 0.93\,/\,0.90 & 0.97\,/\,\textbf{0.50} \\
$5.0$ turns    & Llama-3.1-8B  & 1.00 & 0.15 & 0.95\,/\,0.90 & 0.82\,/\,\textbf{0.10} \\
               & Qwen3-14B     & 1.00 & 0.10 & 0.90\,/\,0.85 & 1.00\,/\,0.70 \\
               & Qwen3-30B-A3B & \underline{0.14} & 0.05 & 0.93\,/\,0.93 & 0.90\,/\,\underline{0.82} \\
               & Mistral-24B   & 1.00 & 0.15 & 0.97\,/\,0.90 & 1.00\,/\,\textbf{0.38} \\
               & Qwen3-32B     & 1.00 & 0.12 & 0.95\,/\,0.93 & 1.00\,/\,0.72 \\
\cmidrule(lr){2-6}
               & \emph{mean}   & 0.84 & 0.11 & 0.94\,/\,0.90 & 0.95\,/\,\textbf{0.54} \\
\bottomrule
\end{tabular}
\caption{\textbf{Plain (undenoised) direction on the full multi-turn sets.} Causal online monitor on cumulative drift; \tool{} is the attributed reminder. catch$_{\mathrm{succ}}$ is the fraction of \emph{successful} attacks flagged. Outcome columns are without\,/\,with mitigation on the same conversations. \underline{Underlined}: Qwen3-30B-A3B, which fails on ActorBreaker.}
\label{tab:mturn}
\end{table}

\section{Scope probes: non-injection detection and intrinsic misuse}
\label{sec:appendix-scope}
These two probes bound where the transition signal applies. They are run on the two attack shapes the main text evaluates---composition across permissible turns, and indirect prompt injection---and ask what happens outside them.

\subsection{Scope: what this signal does and does not read}
\label{sec:rq3}
This paper scopes its claims to two attack shapes: \textbf{multi-turn task decomposition}, where a
harmful goal is assembled from individually permissible user turns (MT-AgentRisk, ActorBreaker), and
\textbf{indirect prompt injection}, where an attacker-controlled instruction enters the agent's context
through a tool return (ASEval, AgentDojo, InjecAgent). It scopes \emph{out} agent-intrinsic misuse,
where the unsafe act is the agent's own decision in response to a directly malicious request with no
foreign instruction and no composition. Two probes mark the edges of that scope. Applied to R-Judge---labeled safe/unsafe trajectories with \emph{no} injection wrapper, re-encoded and scored by max transition drift under leave-one-domain-out CV---the same method reaches $0.77$--$0.80$ AUROC (strong on \textsc{Finance} $\sim\!0.95$, noisy on small-$n$ \textsc{IoT}/\textsc{Web}; full table in Table~\ref{tab:rjudge}), evidence that harmful context updates leave a linear trace even outside prompt injection.

\begin{figure}[t]
\centering
\includegraphics[width=0.62\textwidth]{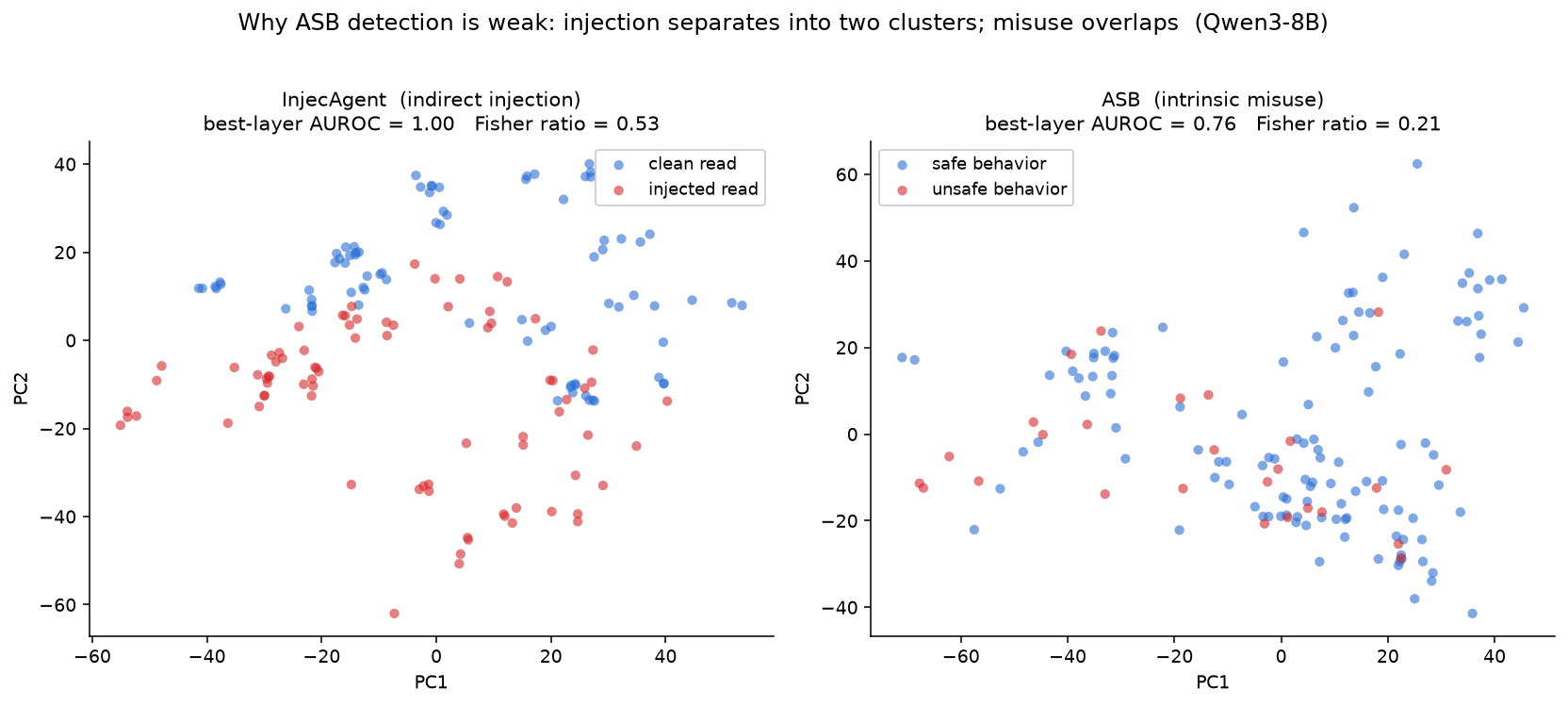}
\caption{Why injection is detectable and intrinsic misuse is not. PCA of transition representations (Qwen3-8B). \emph{Left:} InjecAgent injected (red) vs.\ clean (blue) reads form two clusters (best-layer AUROC $1.00$, Fisher ratio $0.53$). \emph{Right:} ASB unsafe (red) vs.\ safe (blue) \emph{behavior} interleaves (AUROC $0.76$, Fisher $0.21$). Both AUROCs here are best-layer and in-sample, chosen so the panels are comparable; the held-out cross-validated ASB figure is $0.65$--$0.71$ (text). Injection is an external event on a dominant variance axis; misuse is an internal decision leaving a weak, entangled trace.}
\label{fig:pca}
\end{figure}

A second beyond-injection benchmark sharpens the boundary. On \textbf{Agent-SafetyBench} (ASB)~\citep{zhang2024agentsafetybench}---agent \emph{intrinsic} misuse, where the unsafe act is the agent's own tool call in context (e.g.\ externally sharing a confidential file) rather than a foreign instruction---the same transition monitor is markedly weaker: with behavior labels from the benchmark's safe/unsafe judge and a direction re-fit on those labels (matched CV, $\sim$30 unsafe cases/model), held-out AUROC tops out at $0.65$--$0.71$ for the capable models and is near chance for the weakest, and a richer, non-linear classifier over aggregated multi-layer representations does \emph{not} beat difference-of-means. (The $0.76$ quoted in Figure~\ref{fig:pca} is a different quantity: the \emph{best-layer, in-sample} separation of the points that figure plots, reported so that the two panels are comparable to each other. The $0.65$--$0.71$ range is the cross-validated held-out number, and is the one to read as performance.) Figure~\ref{fig:pca} shows why. Projected onto their top principal components, injected-vs-clean read transitions form \emph{two separated clusters} (best-layer AUROC $1.00$, Fisher ratio $0.53$), whereas unsafe-vs-safe \emph{behavior} interleaves (AUROC $0.76$, Fisher $0.21$). Injection is an \emph{external} event: foreign content entering the context restructures the representation enough to become a dominant axis of variation---linearly separable, and detected by every capable model. Intrinsic misuse is an \emph{internal} decision---the offending tool call looks representationally like an ordinary one---so it leaves only a weak, entangled trace, encoded (where at all) in late layers and varying across models. This refines the finding of \S\ref{sec:rq4} that the direction is \emph{primarily structural}: it reads injection-style drift strongly and agent-intrinsic misuse only weakly, and no linear estimator closes the gap---a property of the signal, not the fit. We therefore treat intrinsic misuse (ASB) as a \emph{boundary} result rather than a target of the runtime pipeline. Multi-turn decomposition is different: it is detected near-perfectly once the aggregator is switched to cumulative drift $(h_t-h_0)^{\top}u$, and is evaluated end-to-end in the main text (\S\ref{sec:main_eval}). What it lacks is not detectability but a \emph{correctly attributable} segment---there is no foreign span, and the crossing turn is one of the user's own requests---which is why span redaction does not apply to it, and why read-level withholding, though mechanically applicable, does not suppress the attack: removing the crossing user turn removes a request the user actually made, and the remaining turns still carry the decomposition.

\paragraph{On directly malicious agent benchmarks.} AgentHarm~\citep{andriushchenko2024agentharm}, which issues explicitly malicious multi-step agent tasks across eleven harm categories, sits on the intrinsic-misuse side of this boundary rather than the composition side: the request is harmful in its first turn, nothing is injected, and nothing is assembled from permissible parts. We therefore expect it to behave like Agent-SafetyBench rather than like MT-AgentRisk, and we would predict the transition signal to be weak there for the reason Figure~\ref{fig:pca} gives---the offending tool call is representationally an ordinary one. We have not run it, and we flag it as the natural test of that prediction; a defense of this kind is complementary to refusal training rather than a substitute for it on such inputs.

\begin{table}[t]
\centering\small
\begin{tabular}{lcccccc}
\toprule
\textbf{Model} & App. & Fin. & IoT & Prog. & Web & \textbf{Mean} \\
\midrule
Qwen3-8B  & 0.81 & 0.94 & 0.65 & 0.78 & 0.69 & 0.77 \\
Qwen3-32B & 0.82 & 0.96 & 0.61 & 0.87 & 0.74 & \textbf{0.80} \\
\bottomrule
\end{tabular}
\caption{R-Judge leave-one-domain-out AUROC (non-injection, transition form) over the benchmark's five domains: App.\ (applications and productivity software), Fin.\ (finance), IoT (connected devices), Prog.\ (programming and software engineering), and Web (web browsing and search). The monitor generalizes beyond prompt injection; small-$n$ domains (IoT, Web) are noisier.}
\label{tab:rjudge}
\end{table}

\section{Ablations: aggregator, direction, and steering}
\label{sec:appendix-ablation}
\label{sec:appendix-ablation-main}
\label{sec:ablation-summary}
Three ablations support the design; we summarise them here and give the detail in the subsections that follow. \textbf{(i) Transition vs.\ state:} scoring the transition $\delta_i$ beats the accumulated state $h_i$ on average ($0.925$ vs.\ $0.884$ leave-one-suite-out) and stabilizes the fragile folds, justifying it as the default. \textbf{(ii) What the direction encodes:} controlled probes show it fires on \emph{any} injected imperative---a benign-goal injection in the same wrapper is flagged $98\%$ of the time---and is not a length detector, so it is \emph{primarily structural}, with a weaker but real harm-correlated component. \textbf{(iii) Reader, and deliberately not an actuator:} a one-line reminder dominates every \emph{soft-suffix} steering variant, and a \emph{random}-direction suffix matches steering along $u$, so the suffix's effect is token perturbation rather than displacement along the direction. Direct residual-stream steering \emph{does} do causal work---at a hand-tuned mid layer under strong over-steering it reaches $0.68$ clean flip, above the reminder's $0.52$---but it is confined to that window, adds two uncalibrated hyper-parameters beside a conformally calibrated threshold, discards the attribution, and leaves no auditable record. We therefore detect with $u$ and act symbolically on the flagged context.

The three subsections take these in turn: the aggregator and its representation (transition vs.\ state, \S\ref{sec:rq2}); what the pooled direction actually encodes---and hence whether one general direction suffices (\S\ref{sec:rq4}); and the mitigation lever---reminder vs.\ white-box steering (\S\ref{sec:rq5}).

\subsection{Transition vs.\ state}
\label{sec:rq2}
\tool{} scores the \emph{transition} $\delta_i=h_{i+1}-h_i$ (the indexing of \S\ref{sec:problem-definition}, used throughout)---what the newly appended tool result \emph{added}---rather than the accumulated state $h_i$. Table~\ref{tab:ablation} ablates the two under the identical pipeline. The transition wins on average ($0.925$ vs.\ $0.884$) and never meaningfully loses; the gain is largest exactly where the state form was weakest---the thin-data Mistral ($0.765\!\to\!0.847$), 8B ($0.850\!\to\!0.917$), 32B ($0.886\!\to\!0.932$)---and vanishes at the ceiling (14B). Beyond the mean, the transition \emph{stabilizes} the fragile folds: in the state form the \textsc{travel} held-out fold collapses for several models ($0.55$--$0.77$), because the inner-split layer selector locks onto a spurious early layer; in the transition form those same folds recover to $0.84$--$1.00$. This matches the geometry: differencing cancels the shared baseline---including the depth/length-dependent norm inflation of $h_i$---isolating the component the segment actually contributed. It justifies the transition as the method's default and reduces the whole-state to a baseline. The same transition score, aggregated by \emph{sum} rather than $\max$, is what the deployed monitor uses (\S\ref{sec:runtime-enforcement}): $\max$ suits a single sharp injected segment, the cumulative sum suits harm that accrues over several benign-looking turns.

\begin{table}[t]
\centering\small
\begin{tabular}{lcc}
\toprule
\textbf{Model} & \textbf{State} $h_i$ & \textbf{Transition} $\delta_i$ \\
\midrule
Qwen3-8B        & 0.850 & \textbf{0.917} \\
Qwen3-14B       & \textbf{0.962} & 0.961 \\
Qwen3-32B       & 0.886 & \textbf{0.932} \\
Qwen3-30B-A3B   & 0.900 & \textbf{0.937} \\
Llama-3.1-8B    & 0.939 & \textbf{0.956} \\
Mistral-Small-24B & 0.765 & \textbf{0.847} \\
\midrule
Average         & 0.884 & \textbf{0.925} \\
\bottomrule
\end{tabular}
\caption{Mean leave-one-suite-out AUROC on AgentDojo, held out over its four tool suites (\textsc{workspace}, \textsc{banking}, \textsc{slack}, \textsc{travel}), state vs.\ transition (identical CV, layer $0$ excluded). The transition is the stronger default and lifts the weaker models toward the ceiling.}
\label{tab:ablation}
\end{table}

\subsection{What does the direction encode?}
\label{sec:rq4}
High cross-domain transfer could reflect \emph{structure} (an out-of-place imperative block) rather than harm. We probe this on constructed, matched contexts (Qwen3-8B, $\ell{=}21$). First, the direction \emph{does} transfer across attack \emph{phrasings}: fit on \texttt{important\_instructions}, it detects the unseen \texttt{ignore\_previous} style at $0.96$ accuracy. But a \textbf{benign-long control} exposes what it keys on (Table~\ref{tab:benign}): a long \emph{benign} narrative is never flagged (so it is not a length detector), yet a \emph{benign imperative}---the exact injection wrapper carrying a harmless goal---is flagged $98\%$ of the time, nearly as often as a malicious one. At the operating threshold the direction thus detects \emph{``an injected instruction is present,''} not harmfulness.

A residual harm signal nonetheless exists. Holding the wrapper fixed and varying only the goal, split \emph{by goal} so held-out goals are unseen, malicious vs.\ benign goals still separate at up to $1.00$ (mean $0.97$ over $\ell\in[20,32]$); even against \emph{action-matched} benign goals (harmless but externally side-effecting) separation holds at $0.89$. So the recovered axis is \emph{primarily} injected-instruction structure with a \emph{weaker but real} harm-correlated component.

\begin{table}[t]
\centering\small
\begin{tabular}{lcc}
\toprule
\textbf{Variant (Qwen3-8B, $\ell{=}21$)} & \textbf{Score} & \textbf{Flagged} \\
\midrule
Clean                            & $-14.5$ & $0\%$ \\
Benign, long (no imperative)     & $-7.3$  & $0\%$ \\
Benign imperative (same wrapper) & $+8.6$  & $98\%$ \\
Malicious                        & $+21.5$ & $100\%$ \\
\bottomrule
\end{tabular}
\caption{The direction fires on \emph{any} injected imperative, harmful or not (row 3), and is not driven by length (row 2)---primarily a structural, not semantic-safety, signal.}
\label{tab:benign}
\end{table}

\subsection{Mitigation lever: reminder vs.\ steering}
\label{sec:rq5}
\S\ref{sec:signal} establishes the direction as a \emph{monitor}; we now ask how to \emph{act} on a flag---the enforcement axis of \S\ref{sec:runtime-enforcement}. Comparing mitigations on the agents that actually complied, we find that a lightweight natural-language \textbf{reminder} (\S\ref{sec:runtime-enforcement}) is the most effective lever---and (below) cheap enough to apply always-on---whereas the tempting white-box alternative---steering along the reading direction $u$ itself, a recurring promise of representation engineering~\citep{zou2023representation,rimsky2024steering}---is dominated by it and, we show, largely \emph{correlational rather than causal}. The reminder is thus our deployable mitigation; the steering study explains \emph{why} the simpler lever is the right one.

\paragraph{Protocol.} From the real rollouts we take every trace in which the agent \emph{executed} the injected action ($111$ complying trajectories over the three highest-compliance models---Qwen3-14B/32B/30B-A3B---and all four suites). For each we reconstruct the pre-action context, intervene, and greedily regenerate. Success is \textbf{clean flip}: the attacker action absent \emph{and} the response still coherently pursues the user task (a legitimate tool call for the action suites; a non-attacker recommendation for \textsc{travel}). The steering direction is fit \emph{leave-one-suite-out}, exactly as the monitor is, so the enforcer never sees its target domain. We compare, at a matched operating point, four interventions. The \textbf{soft suffix} is a
continuous prefix-tuning baseline, defined here and \emph{not} part of \tool{}: $k{=}8$ free embedding
vectors $e_{1:k}$ are appended to the pre-action context and optimised by Adam (lr $10^{-2}$, $200$
steps) to minimise the negative log-likelihood of a fixed refusal continuation while maximising the
projection gap $-(z^{\top}u)$ at the read layer, under the norm budget
$\lVert e_j\rVert_2\le\lVert\bar e\rVert_2$ where $\bar e$ is the mean token embedding; the same
optimiser, budget and step count are used for every variant. We run it with a leave-one-suite-out
(\textsc{loso}) direction and with an \emph{in-domain} direction, and against a \textbf{random}-direction
suffix (identical optimiser and budget, $u$ replaced by a random unit vector) which isolates how much
of any effect is the appended tokens rather than the direction. The fourth is the deployed one-line
defensive \textbf{reminder} appended as a user turn (\S\ref{sec:runtime-enforcement}).

\paragraph{The reminder is the effective mitigation.} Table~\ref{tab:enforce} reports clean flip. The reminder \emph{dominates every model} ($0.20$--$0.52$); the leave-one-out soft suffix redirects almost never ($\le 0.10$) and the in-domain direction is barely better. Tellingly, the \textbf{random} direction matches or beats the injection direction on the two larger models, so the little suppression that occurs is not coming from $u$. A layer sweep \emph{of the soft suffix} confirms the confound: the injected-vs-clean separation along $u$ grows $\sim\!500\times$ with depth (from $0.3$ to $160$ projection units; the ``sep.'' column of Table~\ref{tab:steerlayer}), yet the suffix's clean-flip rate is flat across that whole range ($0.00$--$0.33$, no depth trend), redirecting as often at frac $0.2$, where the direction \emph{barely exists}, as at frac $0.65$, where it is strongest. The soft suffix's effect is the appended tokens \emph{perturbing} generation, not displacement along $u$.

\paragraph{Always-on vs.\ gated.} The only difference between an always-on and a monitor-gated reminder is whether \emph{clean} turns are reminded, so we measure the cost of the always-on policy directly. On $32$ clean pre-action contexts (Qwen3-14B) the reminder preserves the agent's tool call in $100\%$ of cases and the exact target function in $94\%$: an always-on reminder is \emph{clean-task-safe}, and a strong, simple baseline. Gating it on the monitor therefore only saves the reminder on unflagged turns---an optional refinement, not a utility necessity. The monitor's contribution is \emph{detection}; the reminder is a cheap mitigation to pair with it.

\paragraph{Direct steering: real control, in a window that must be known in advance.} Removing the
perturbation confound---adding
\[
-\alpha\bigl(z^{\top}u-\mu^{-}\bigr)\,u
\]
into the residual stream at the flagged layer during generation, with no appended tokens---the
direction does genuine causal work. Here $z\in\mathbb{R}^{d}$ is the residual-stream activation at the
current token and the flagged layer, $u$ is the unit reading direction, $\alpha$ is the steering
coefficient, and $\mu^{-}=\mathbb{E}[z^{\top}u\mid\text{clean}]$ is the mean projection of
\emph{clean} calibration activations along $u$, estimated on the fitting split. The edit thus pulls the
current activation's projection back toward its clean mean. (We write $\mu^{-}$ rather than
$\rho$ to avoid collision with the reminder template $\rho(c^\star)$ of
\S\ref{sec:runtime-enforcement}.) We replicate the sweep at $6.2\times$ the original sample ($37$ complying Qwen3-14B cases across all four suites): clean flip is confined to \emph{middle} layers under \emph{strong over-steering}, $48/111$ ($0.43$) at $\alpha{=}2$ in the frac $0.35$--$0.65$ band against $5/111$ ($0.05$) outside it and $0/222$ unsteered (Fisher $p<10^{-3}$), and the peak cell reproduces the small-sample estimate almost exactly ($0.68$ at frac $0.5$, against $0.67$ at $n{=}6$). Gentle steering stays weak ($\le\!0.16$ at $\alpha\!\le\!1$) and late-layer steering all but fails ($0.02$ at frac $0.8$, $0.04$ at frac $0.9$) even though it displaces the projection furthest---by then the action is decided and steering only corrupts the surface form. At its optimum steering is therefore \emph{stronger} than the one-line reminder ($0.68$ vs.\ $0.52$, same model and suites), but that optimum is a hand-tuned layer and over-steer coefficient with no calibration procedure, and it collapses by an order of magnitude one band away.

\paragraph{Why we act on the context rather than on the model.} Efficacy is not why \tool{} declines to steer, and we are explicit that on the injected-run trade-off steering can win. The reason is architectural. An edit to the residual stream is blind to \emph{which} segment fired: it discards the attribution the monitor has just computed, and it leaves no trace in the transcript, so nothing records that enforcement occurred, on what evidence, or to what effect. A quoted span (the deployed reminder, \S\ref{sec:runtime-enforcement}) or a withheld segment (the read-level fallback of Appendix~\ref{sec:appendix-fallback}) is inspectable, reviewable, and reversible; a steering edit is none of these, and it cannot supply the provenance on which either enforcement lever depends. Steering also adds two uncalibrated hyper-parameters---layer and $\alpha$---beside a threshold that carries a conformal false-alarm guarantee. This is consistent with what $u$ is: difference-of-means maximizes \emph{correlation} with the injected/clean label (\S\ref{sec:rq4}: largely structural), which yields an excellent readout with no guarantee that displacing along it reconstructs correct behavior~\citep{belinkov2022probing,elazar2021amnesic}, even where---as here---it demonstrably moves the action.

\begin{table}[t]
\centering\small
\begin{tabular}{lcccc}
\toprule
\textbf{Model} & \textsc{loso} & in-dom.\ & random & \textbf{reminder} \\
\midrule
Qwen3-14B           & 0.10 & 0.24 & 0.05 & \textbf{0.52} \\
Qwen3-32B           & 0.05 & 0.05 & 0.05 & \textbf{0.47} \\
Qwen3-30B-A3B (MoE) & 0.00 & 0.07 & 0.13 & \textbf{0.20} \\
\bottomrule
\end{tabular}
\caption{Clean-flip rate (injected action suppressed \emph{and} user task preserved) at the operating point, per intervention. The white-box soft suffix (\textsc{loso}/in-domain) is dominated by a one-line prompt \emph{reminder}, and does not beat a \emph{random}-direction suffix on the larger models---the suppression is not coming from the direction.}
\label{tab:enforce}
\end{table}

\begin{table}[t]
\centering\small
\begin{tabular}{lccccc}
\toprule
\textbf{layer} & \textbf{sep.} & $\alpha{=}0$ & $\alpha{=}0.5$ & $\alpha{=}1$ & $\alpha{=}2$ \\
\midrule
frac 0.20 & 0.3 & 0.00 & 0.00 & 0.00 & 0.00 \\
frac 0.35 & 14  & 0.00 & 0.00 & 0.00 & 0.50 \\
frac 0.50 & 20  & 0.00 & 0.00 & 0.17 & \textbf{0.67} \\
frac 0.65 & 61  & 0.00 & 0.00 & 0.00 & 0.50 \\
frac 0.80 & 99  & 0.00 & 0.00 & 0.00 & 0.00 \\
frac 0.90 & 160 & 0.00 & 0.00 & 0.00 & 0.00 \\
\bottomrule
\end{tabular}
\caption{Direct-steering clean flip by layer depth (fraction) $\times$ steering strength $\alpha$ (Qwen3-14B, pooled \textsc{slack}+\textsc{travel}). ``sep.'' is the injected-vs-clean projection separation, which grows $\sim\!500\times$ with depth. Control appears only in a mid-layer, strong-$\alpha$ window; late layers move the projection most yet flip nothing. Cells here are $n{=}6$; a $6.2\times$ replication ($n{=}37$, all four suites) reproduces the peak cell at $0.68$ and confirms the depth localisation---$0.43$ pooled in the frac $0.35$--$0.65$ band at $\alpha{=}2$ against $0.05$ outside and $0.00$ unsteered, Fisher $p<10^{-3}$.}
\label{tab:steerlayer}
\end{table}

\subsection{What to quote, and what to do when quoting backfires}
\label{sec:appendix-fallback}\label{sec:appendix-attr}

\S\ref{sec:runtime-enforcement} commits to one lever---an attributed natural-language reminder---and
Table~\ref{tab:sturn} shows it inverting on 8B-class models. This subsection separates the two
questions that failure raises: \emph{which} segment the reminder should name, and \emph{what to do
instead} when naming one makes things worse.

\paragraph{Which segment to name.} Table~\ref{tab:attrmode} varies only the quoted content on
MT-AgentRisk, holding the direction, the threshold, the crossing turn and the decoding fixed, so the
arms differ in nothing else. Three rules are compared: an \textbf{unattributed} warning that names no
segment; the single largest contributor $\arg\max_j r_j$; and a \textbf{path-based} top-$2$ rule that
names the two largest contributors in turn order---the cumulative-attribution variant a
distributed attack would call for. Two things follow. First, path-based attribution changes
essentially nothing here: mean ASR $0.444$ against $0.438$, and only $2$ of $160$ harmful
trajectories name a different segment under the two rules at all, because MT-AgentRisk attacks run
$3.2$ turns and many crossings occur when a single candidate exists. Whether a cumulative rule earns
its cost on a genuinely long, thinly-primed attack is not settled by this measurement, and we do not
claim it is. Second, and more interestingly, \emph{whether to quote at all is model-dependent in the
same direction as the inversion}: naming a segment helps the three larger models
($0.650\!\to\!0.525$, $0.425\!\to\!0.325$, $0.625\!\to\!0.500$) and \emph{hurts} Qwen3-8B
($0.300\!\to\!0.400$). The abstract warning is not a free fix---it costs $0.06$ ASR on average---but
it is the better lever on exactly the models where quoting inverts.

\begin{table}[t]
\centering\small
\setlength{\tabcolsep}{6pt}
\begin{tabular}{lcccc}
\toprule
& \textbf{unattributed} & $\arg\max_j r_j$ & \textbf{top-2 (path)} & \textbf{\# discriminating} \\
\textbf{Model} & ASR\,/\,Util & ASR\,/\,Util & ASR\,/\,Util & trajectories \\
\midrule
Qwen3-8B      & \textbf{0.300}\,/\,0.825 & \underline{0.400}\,/\,0.775 & 0.400\,/\,0.775 & 0 \\
Qwen3-14B     & 0.650\,/\,0.850 & \textbf{0.525}\,/\,0.875 & 0.525\,/\,0.875 & 0 \\
Qwen3-30B-A3B & 0.425\,/\,0.825 & \textbf{0.325}\,/\,0.825 & 0.350\,/\,0.825 & 2 \\
Qwen3-32B     & 0.625\,/\,0.850 & \textbf{0.500}\,/\,0.875 & 0.500\,/\,0.875 & 0 \\
\midrule
\emph{mean}   & 0.500 & \textbf{0.438} & 0.444 & 2\,/\,160 \\
\bottomrule
\end{tabular}
\caption{\textbf{What the reminder should quote} (MT-AgentRisk, $40$ harmful and $40$ benign
trajectories per model, identical direction, identical $\tau$, identical crossing turns). ``\#
discriminating'' counts harmful trajectories on which $\arg\max$ and top-$2$ actually name different
content; where the crossing occurs at turn $0$ the two coincide by construction. Path-based
attribution is within noise of the single $\arg\max$. \underline{Underlined}: Qwen3-8B, where quoting
a segment is \emph{worse} than an abstract warning---the same capability gate that inverts the
reminder in Table~\ref{tab:sturn}.}
\label{tab:attrmode}
\end{table}

\paragraph{What to do when quoting backfires.} Table~\ref{tab:fallback} puts the reminder beside two
levers that do not depend on the model reading a quotation correctly. \textbf{Halting} forces a
refusal on a flag. \textbf{Read-level withholding} replaces the attributed segment with a placeholder
noting that a read was withheld, re-anchors the accumulator at the post-removal state, and lets the
agent continue; it needs read-level attribution, which is measured at $1.00$ on these runs, but no
sub-segment localisation. Three findings. \emph{(i) The reminder is capability-gated, not broken}: it
takes ASR to $0.00$--$0.10$ on four of six models and to $0.00$ on both InjecAgent families for the
14B, but rises above the no-intervention baseline on Qwen3-8B ($0.23\!\to\!0.29$ on AgentDojo,
$0.33\!\to\!0.87$ on InjecAgent \textsc{ds}) and Llama-3.1-8B ($0.07\!\to\!0.38$). \emph{(ii)
Withholding is uniformly reliable}: ASR $\le 0.09$ on every model and both benchmarks, including every
cell where the reminder inverts. \emph{(iii) It is not free}: the reminder preserves more task
completion than withholding on five of six models, and halting---which is the most reliable of all at
ASR $0.00$---is also by far the most expensive, leaving $0.00$--$0.24$ completion on the flagged runs
against the reminder's $0.37$--$0.71$.

This gives a deployment rule we can state without tuning: use the attributed reminder where the
model is strong enough to read a quotation as a negative example, and fall back to read-level
withholding where it is not. Which regime a model is in is measurable before deployment, on the same
red-team calibration set that fits the direction---issue the reminder on the calibration attacks and
check whether ASR goes down. We report the two levers side by side rather than folding the choice
into \tool{} itself, because the gate is a property of the model rather than of the monitor.

\begin{table}[t]
\centering\small
\setlength{\tabcolsep}{5pt}
\begin{tabular}{llccccc}
\toprule
& & & \multicolumn{4}{c}{\textbf{ASR\,/\,task completion on the flagged runs}} \\
\cmidrule(lr){4-7}
\textbf{Benchmark} & \textbf{Model} & $n$ & no interv. & halt & \textbf{reminder} & withhold read \\
\midrule
AgentDojo      & Qwen3-8B      & 35 & 0.23\,/\,0.31 & 0.00\,/\,0.14 & \underline{0.29}\,/\,0.37 & \textbf{0.00}\,/\,0.29 \\
\textsc{slack} & Llama-3.1-8B  & 30 & 0.03\,/\,0.43 & 0.00\,/\,0.00 & \textbf{0.00}\,/\,0.50 & 0.03\,/\,0.00 \\
               & Qwen3-14B     & 42 & 0.45\,/\,0.57 & 0.00\,/\,0.12 & \textbf{0.05}\,/\,0.45 & 0.05\,/\,0.24 \\
               & Qwen3-30B-A3B & 21 & 0.33\,/\,0.62 & 0.00\,/\,0.24 & 0.10\,/\,\textbf{0.71} & 0.05\,/\,0.71 \\
               & Mistral-24B   & 35 & 0.71\,/\,0.60 & 0.00\,/\,0.14 & 0.26\,/\,0.60 & \textbf{0.09}\,/\,0.43 \\
               & Qwen3-32B     & 40 & 0.62\,/\,0.45 & 0.00\,/\,0.12 & \textbf{0.00}\,/\,0.38 & 0.03\,/\,0.28 \\
\midrule
InjecAgent     & Qwen3-8B \textsc{dh}  & 15 & 0.14 & 0.00 & \underline{0.29} & \textbf{0.00} \\
($+$enhanced)  & Qwen3-8B \textsc{ds}  & 15 & 0.33 & 0.00 & \underline{0.87} & \textbf{0.00} \\
               & Qwen3-14B \textsc{dh} & 15 & 0.29 & 0.00 & \textbf{0.00} & \textbf{0.00} \\
               & Qwen3-14B \textsc{ds} & 15 & 0.20 & 0.00 & \textbf{0.00} & \textbf{0.00} \\
\bottomrule
\end{tabular}
\caption{\textbf{Fallbacks where the attributed reminder inverts.} Measured on the \emph{flagged}
trajectories only ($n$ per cell), so that every arm acts on the same runs and the comparison isolates
the intervention from the detector. ``halt'' forces a refusal; ``withhold read'' replaces the
attributed segment with a withheld-content placeholder and re-anchors the accumulator.
InjecAgent reports ASR only: its utility criterion is satisfied before the injected segment is read,
so completion is near-saturated by construction and not comparable to AgentDojo's.
\underline{Underlined}: cells where the reminder raises ASR above no intervention. Withholding holds
ASR $\le 0.09$ everywhere, including those cells; halting is the most reliable and the most expensive.}
\label{tab:fallback}
\end{table}

\section{Monitoring overhead}
\label{sec:appendix-overhead}

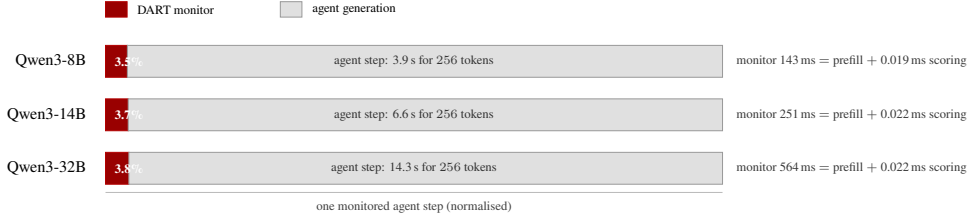
\begin{figure}[t]
\centering
\resizebox{0.92\textwidth}{!}{%
\begin{tikzpicture}[font=\small]
  \node[left,font=\small] at (-0.25,0.31) {Qwen3-8B};
  \draw[fill=red!60!black,draw=red!70!black] (0,0.00) rectangle (0.425,0.62);
  \draw[fill=black!12,draw=black!40] (0.425,0.00) rectangle (12.0,0.62);
  \node[font=\scriptsize,white,right] at (0.06,0.31) {\textbf{3.5\%}};
  \node[font=\scriptsize,gray!40!black] at (6.00,0.31) {agent step: 3.9\,s for $256$ tokens};
  \node[font=\scriptsize,right,gray!45!black] at (12.15,0.31) {monitor 143\,ms $=$ prefill $+$ 0.019\,ms scoring};
  \node[left,font=\small] at (-0.25,-0.73) {Qwen3-14B};
  \draw[fill=red!60!black,draw=red!70!black] (0,-1.04) rectangle (0.442,-0.42);
  \draw[fill=black!12,draw=black!40] (0.442,-1.04) rectangle (12.0,-0.42);
  \node[font=\scriptsize,white,right] at (0.06,-0.73) {\textbf{3.7\%}};
  \node[font=\scriptsize,gray!40!black] at (6.00,-0.73) {agent step: 6.6\,s for $256$ tokens};
  \node[font=\scriptsize,right,gray!45!black] at (12.15,-0.73) {monitor 251\,ms $=$ prefill $+$ 0.022\,ms scoring};
  \node[left,font=\small] at (-0.25,-1.77) {Qwen3-32B};
  \draw[fill=red!60!black,draw=red!70!black] (0,-2.08) rectangle (0.455,-1.46);
  \draw[fill=black!12,draw=black!40] (0.455,-2.08) rectangle (12.0,-1.46);
  \node[font=\scriptsize,white,right] at (0.06,-1.77) {\textbf{3.8\%}};
  \node[font=\scriptsize,gray!40!black] at (6.00,-1.77) {agent step: 14.3\,s for $256$ tokens};
  \node[font=\scriptsize,right,gray!45!black] at (12.15,-1.77) {monitor 564\,ms $=$ prefill $+$ 0.022\,ms scoring};
  \draw[gray!55] (0,-2.24) -- (12.0,-2.24);
  \node[font=\scriptsize,gray!50!black] at (6.00,-2.53) {one monitored agent step (normalised)};
  \draw[fill=red!60!black,draw=red!70!black] (0,1.17) rectangle (0.42,1.47);
  \node[right,font=\scriptsize] at (0.5,1.3199999999999998) {\tool{} monitor};
  \draw[fill=black!12,draw=black!40] (3.4,1.17) rectangle (3.82,1.47);
  \node[right,font=\scriptsize] at (3.9,1.3199999999999998) {agent generation};
\end{tikzpicture}}
\caption{\textbf{Monitoring overhead.} Per monitored segment \tool{} performs one prefill and one dot product against $u$; bars are normalised to one monitored step, absolute times at right. The monitor reuses the agent's own weights, so no second model is resident, and the prefill re-reads states the forward pass has already computed---an integrated implementation pays only the dot product, so this is an upper bound. Median over MT-AgentRisk conversations on one M3 Ultra.}
\label{fig:overhead}
\end{figure}

\paragraph{What is measured.} Per monitored segment, \tool{} performs one forward pass over the
context to read $h_{i+1}$ at the layer $\ell^\star$, and one dot product against $u$. Figure
\ref{fig:overhead} reports the wall-clock cost of both against the cost of the agent step they sit
beside, on one Apple M3 Ultra ($96$\,GB unified memory) under MLX with 8-bit weights, taking
the median over MT-AgentRisk conversations. The agent step is fixed at $256$ generated tokens so the
denominator is stable across models; the monitor's cost is the cost of a prefill over the
\emph{current} context, which grows with the conversation, so the reported figure is the median over
all monitored steps of all conversations rather than a single-length measurement. Contexts in this
setting run to a few thousand tokens by the final turn.

\paragraph{KV-cache and why this is an upper bound.} We measure the monitor as a \emph{separate}
prefill: it re-reads the context through the same weights, with its own cache, and does not reuse the
agent loop's KV-cache. This is the pessimistic implementation, and we report it because it is the one
that requires no change to the serving stack. An integrated implementation would tap $h_{i+1}$ from
the forward pass the agent already runs over the newly appended segment---the state is computed there
regardless---and would pay only the dot product, which is $0.019$--$0.022$\,ms, four
orders of magnitude below the prefill. The figures in Figure~\ref{fig:overhead} should therefore be
read as a ceiling on the integration cost, not as an intrinsic one.

\paragraph{Memory.} The monitor adds no resident model: it reuses the agent's own weights, so the only
additional state is the reading direction $u$ (one vector of the model's hidden width: $4096$ floats on
Qwen3-8B, $5120$ on Qwen3-32B, i.e.\ tens of kilobytes) and the transient activations of the extra
prefill, which are released immediately. Peak memory over the agent's own requirement was within
measurement noise in our runs. This is the cost line on which \tool{} differs most sharply from
speculative simulation (\S\ref{sec:baselines}), which must hold a second model resident and run it
several times per step.

\end{document}